\documentclass[lettersize,journal]{IEEEtran}
\usepackage{amsmath,amsfonts}
\usepackage{algorithmic}
\usepackage{algorithm}
\usepackage{array}
\usepackage[caption=false,font=normalsize,labelfont=sf,textfont=sf]{subfig}
\usepackage{textcomp}
\usepackage{stfloats}
\usepackage{url}
\usepackage{verbatim}
\usepackage{graphicx}
\usepackage{cite}
\usepackage{multicol}
\usepackage{multirow}
\usepackage{color}
\usepackage{xspace}
\usepackage{epstopdf}
\usepackage{url}
\usepackage{breqn}
\usepackage{booktabs}
\usepackage[table,xcdraw]{xcolor}
\usepackage{xcolor,colortbl}
\usepackage{float}

\newcommand{\wx}[1]{\textcolor{black}{#1}}
\newcommand{\qi}[1]{\textcolor{black}{#1}}

\newcommand{\etal}{\textit{et al.}}

\begin{document}

\title{Memory-Constrained Semantic Segmentation for Ultra-High Resolution UAV Imagery}

\author{Qi Li,
        Jiaxin Cai,
        Yuanlong Yu,
        Jason Gu,
        Jia Pan,
        Wenxi Liu}


\markboth{Journal of \LaTeX\ Class Files,~Vol.~14, No.~8, August~2021}%
{Shell \MakeLowercase{\textit{et al.}}: A Sample Article Using IEEEtran.cls for IEEE Journals}


\maketitle

\begin{abstract}
Amidst the swift advancements in photography and sensor technologies, high-definition cameras have become commonplace in the deployment of Unmanned Aerial Vehicles (UAVs) for diverse operational purposes.
Within the domain of UAV imagery analysis, the segmentation of ultra-high resolution images emerges as a substantial and intricate challenge, especially when grappling with the constraints imposed by GPU memory-restricted computational devices.
This paper delves into the intricate problem of achieving efficient and effective segmentation of ultra-high resolution UAV imagery, while operating under stringent GPU memory limitation.
The strategy of existing approaches is to  downscale the images to achieve computationally efficient segmentation. However, this strategy tends to overlook smaller, thinner, and curvilinear regions. 
To address this problem, we propose a GPU memory-efficient and effective framework for local inference without accessing the context beyond local patches. In particular, we introduce a novel spatial-guided high-resolution query module, which predicts pixel-wise segmentation results with high quality only by querying nearest latent embeddings with the guidance of high-resolution information. Additionally, we present an efficient memory-based interaction scheme to correct potential semantic bias of the underlying high-resolution information by associating cross-image contextual semantics. For evaluation of our approach, we perform comprehensive experiments over public benchmarks and achieve superior performance under both conditions of small and large GPU memory usage limitations.
\end{abstract}


\section{Introduction}

\IEEEPARstart{W}{ith} the rapid progress of photography and sensor technologies, high-definition cameras have become commonplace in the deployment of Unmanned Aerial Vehicles (UAVs).
Thus, there is a growing demand for ultra-high resolution (i.e., 2K, 4K, or even higher resolution) of UAV imagery for diverse applications, such as UAV localization \cite{goforth2019gps}, UAV detection \cite{zheng2021air}, and agricultural monitoring \cite{valente2019fast, li2016large}. 

However, handling ultra-high resolution images will cost unaffordable computing resources, which is a formidable challenge for robotic systems with limited computation power~\cite{ferrarini2022binary, grossman2023just, gomez2020real, hu2020real, sun2021lightweight}. 
The strategy of existing UAV imagery analysis approaches is to first downscale the ultra-high resolution images to achieve computationally efficient segmentation, but this strategy tends to overlook smaller, thinner, and curvilinear regions. 
More importantly, as input image resolutions continue to increase, this method demands a substantially greater amount of GPU memory, rendering it impractical for systems with limited GPU memory resources.
Thus, is it possible to handle semantic segmentation of arbitrarily large images using limited GPU memory? 

In this paper, we attempt to address such a new  ultra-high resolution segmentation problem under limited GPU memory constraints. 
To resolve this problem, it is appropriate to resort to segmenting local patches instead of the entire ultra-high resolution image before merging all local segmentation results. However, this approach often leads to degraded performance, and thus additional context beyond the local patch needs to be introduced to address this concern in prior methods~\cite{chen2019collaborative, huynh2021progressive, li2021contexts}. Unfortunately, the introduction of additional context also results in increased GPU memory usage.
In this paper, we propose a novel method for performing local segmentation without relying on the context beyond local patches. This approach results in a GPU memory-efficient and effective framework of ultra-high resolution semantic segmentation.


In particular, we draw inspiration from the idea of implicit neural representation (INR)~\cite{chen2021learning} and design an efficient spatial-guided high-resolution query module,
enabling our model to infer high-quality pixel-wise segmentation results. In specific, our model queries the nearest latent embeddings of the spatial coordinates and the high-resolution spatial information as guidance, reducing the dependency on extra contextual information beyond the local patch to the largest extent. 
Moreover, we propose to guide the latent embeddings to supplement the details through high-resolution semantic masks in a more straightforward manner. However, the high-resolution spatial information tends to introduce semantic estimation bias during inference. \qi{To address this concern, we introduce a memory-based interaction scheme that efficiently facilitates the high-resolution semantics learning from compact cross-image contextual representation.
Compared with previous memory-based schemes, our designed scheme adds only 1MB of GPU memory computational overhead thanks to its linear complexity.}

Through comprehensive experiments, we demonstrate that our proposed model outperforms the state-of-the-art approaches under the condition of small GPU memory limits over Inria Aerial and DeepGlobe datasets. Besides, our model also can better trade-off segmentation performance, GPU memory usage, and computational overhead than the latest off-the-shelf methods in large GPU memory-limited systems.

Overall, the main contributions of our paper are:
\begin{itemize}
    \item In this paper, we raise a new research problem on ultra-high image segmentation under a GPU memory-constrained condition. We propose a GPU memory-efficient and effective framework to handle such a challenging problem.
    \item A novel spatial-guided high-resolution query module is introduced to predict semantic masks of local image patches without requiring additional contextual cues beyond the local region.
    \item We propose an efficient memory-based interaction scheme to address the issue of semantic bias arising from the local nature of image patches, which incorporates cross-image contextual information for high-resolution query, and introduces only a mild extra GPU memory overhead. 
    \item Our model achieves superior performance in ultra-high resolution image segmentation, surpassing prior methods by a substantial margin, particularly under small GPU memory-limited conditions. Moreover, our approach offers a balanced trade-off between segmentation performance, GPU memory usage, and computational overhead under large GPU memory constraints.
\end{itemize}

\section{Related Works}

In this section, we survey the recent progress of ultra-high resolution semantic segmentation and introduce the related literatures on implicit neural representation and memory schemes.

\subsection{Ultra-high Resolution Semantic Segmentation}

Semantic segmentation is modeled as a dense prediction task, many works \cite{simonyan2014very, long2015fully, he2016deep, zhao2017pyramid, badrinarayanan2017segnet, chen2018encoder, wang2020deep} based on convolutional neural network has achieved great success. In recent years, several methods \cite{zheng2021rethinking, liu2021swin, xie2021segformer, strudel2021segmenter, cheng2021per, cheng2022masked} use Transformer architecture to conduct semantic segmentation tasks. However, most of the work is applied to ultra-high resolution images, which raises the trade-off between performance and GPU memory. To this end, Chen \etal \cite{chen2019collaborative} integrate global image and local patch each other in the deep layer to balance performance and GPU memory usage. Limited by the speed of global and local interaction, Wu \etal \cite{wu2020patch} design a classification network to choose important patches for the feature fusion. To further improve performance, Huynh \etal \cite{huynh2021progressive} progressively refine coarse segmentation results via a multi-stage processing pipeline. Li \etal \cite{li2021contexts} introduce a multi-scale locality-aware contextual correlation and the adaptive feature fusion scheme to strengthen local segmentation. These methods are patch-based approaches, which can save GPU memory but consume time due to multiple local segmentation. In the latest work, Guo \etal \cite{guo2022isdnet} leverage the shallow-deep network to directly process the full-scale ultra-high resolution images for accelerating the inference speed. In addition, some work \cite{cheng2020cascadepsp, shen2022high} generates high-quality semantic results by refining coarse segmentation maps from a pre-trained model. Comparing with the previous works, we focus on the systems constrained by limited GPU memory, by considering a better trade-off between accuracy, GPU memory, and speed.

\subsection{Implicit Neural Representation}

In implicit neural representation (INR), the signals of the object and scene are maps from coordinates via a multi-layer perceptron (MLP) applied in modeling 3D reconstruction \cite{park2019deepsdf, sitzmann2019scene, chen2019learning, niemeyer2020differentiable,mildenhall2020nerf,jiang2020local}. For example, Mildenhall \etal \cite{mildenhall2020nerf} present NeRF that learns an implicit representation for a novel scene view using multiple image views. It can accurately capture the intricate details of the shape with a minimal amount of parameters. Later, INR is also used in video representation \cite{chen2021nerv, li2022nerv, shangguan2022learning}. Chen \etal \cite{chen2021nerv} propose a novel neural representation for videos that takes the time index as input and directly outputs the corresponding RGB frame.
Recently, INR has also made some progress in 2D tasks \cite{chen2021learning, xu2021ultrasr, shen2022high, hu2022learning}. Chen \etal \cite{chen2021learning} represent natural and complex images in a continuous manner, which are trained in the image super-resolution task. Xu \etal \cite{xu2021ultrasr} perform the spatial encoding in implicit functions and further introduce deep coordinate fusion and residual MLP architecture. Hu \etal \cite{hu2022learning} propose an alignment function using multi-level feature fusion for semantic segmentation. Among them, \cite{hu2022learning} is most related to our work. The key difference rests in that we utilize high-resolution spatial masks as guidance to the query module for the interpretable details of the ultra-high resolution images, while \cite{hu2022learning} only utilizes INR to perform multi-scale feature alignment.

\subsection{Memory Scheme}

Similar to the human brain, deep neural networks encode, store, and extract information via memory. In the vision tasks \cite{oh2019video, wang2020cross, chen2020memory, hu2021learning, xie2021efficient, kim2022pin, jin2022mcibi++}, it can capture cross-image information to serve the current image. Wang \etal \cite{wang2020cross} use a cross-batch memory mechanism to record and update the embeddings of past iterations for the collection of sufficient hard negative pairs. Xie \etal \cite{xie2021efficient} relieve the ambiguity of similar objects by memorizing and tracking the regions of target objects. In \cite{kim2022pin}, Kim \etal store the domain-agnostic categorical knowledge in the memory to achieve domain generalization for semantic segmentation. Jin \etal \cite{jin2022mcibi++} set up a memory module to store the dataset-level distribution information of all classes and perform a coarse-to-fine iterative inference strategy in the memory. In our work, we introduce a memory-based interaction scheme that stores low-resolution semantic information to efficiently enhance the spatial semantics of the high-resolution image. Therefore, it is able to rectify spatial-guided semantic bias in our query module.

\begin{figure*}
    \centering
    \includegraphics[width=0.95\textwidth]{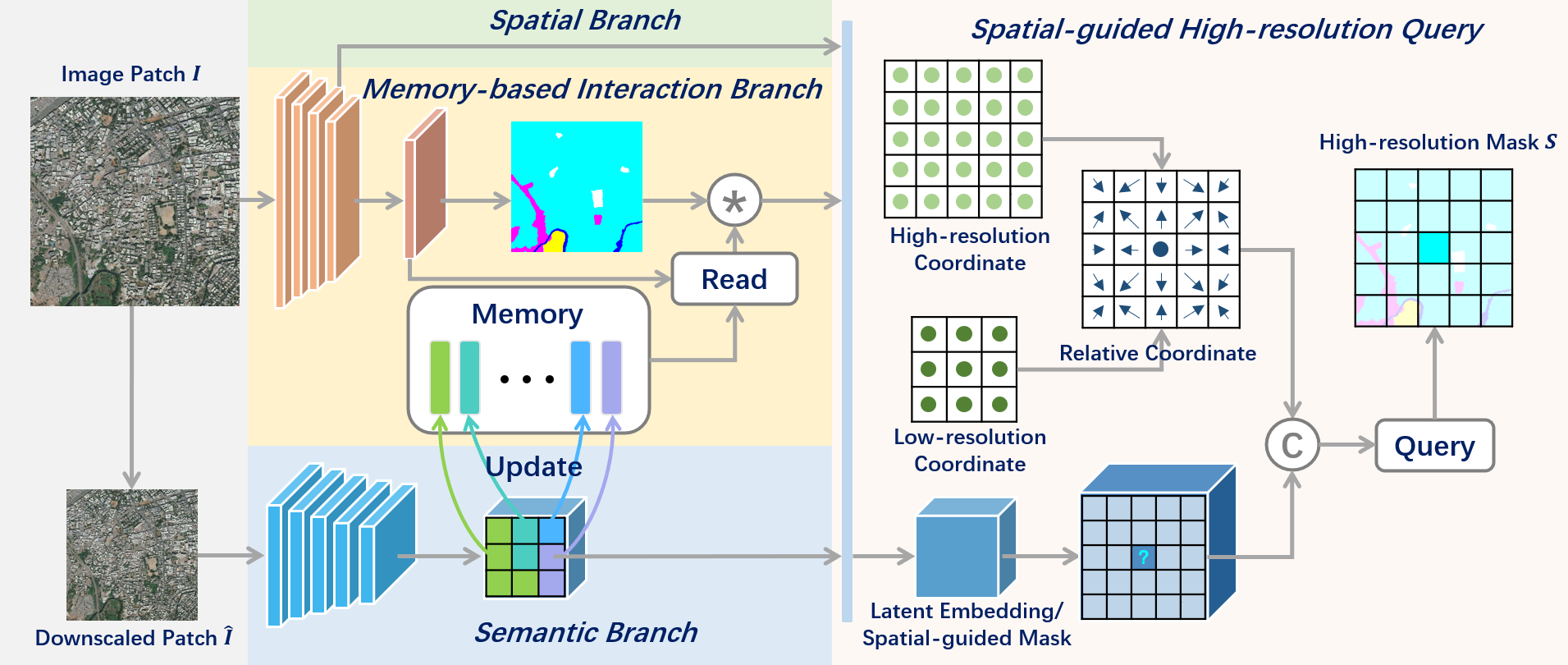}
    \caption{The pipeline of our model. In specific, the input image is passed into a three-branch architecture to extract the low-resolution latent embedding and high-resolution spatial mask. They are jointly input to the spatial-guided high-resolution query module to obtain high-quality segmentation results. Besides, the memory-based interaction scheme is introduced to correct the semantic deviation of guide information.}
    \label{fig:framework}
\end{figure*}

\section{Methodology}

In this section, we first describe the overall pipeline of our framework. Then, we elucidate the core components of our model, i.e., spatial-guided high-resolution query module and  memory-based interaction scheme.

\subsection{Framework Overview}

In this paper, we introduce a semantic segmentation approach capable of efficiently processing ultra-high resolution images on systems with limited GPU memory. 

\wx{In response to the GPU memory constraint, we follow the patch-based paradigm (e.g., \cite{li2021contexts}) in which we partition a large image into local patches, segment them subsequently, and then merge all the local segmentation results together.} Formally, given an ultra-high resolution image $\dot{I}$ with width $\dot{W}$ and height $\dot{H}$, our approach partitions the image into $N$ overlapping local patches (denoted as $I$ in the following) along both the row and column axes. Then, our model generates the segmentation results for each local patch, which are merged to produce an ultra-high resolution segmentation mask.

Given the possibly large resolution of the image patch $I$ (\wx{e.g. 1280 $\times$ 1280}), there still arises a necessity to mitigate GPU memory expenses. This, in turn, imposes constraints on the depth of the encoding network, which consequently limits its capability to extract an ample amount of semantic information. Therefore, we downscale $I$ to a properly small resolution (\wx{e.g. 320 $\times$ 320}) for extracting semantics. 
To this end, as illustrated in Fig.~\ref{fig:framework}, our proposed framework is based on a three-branch architecture, comprising of the semantic branch, memory-based interaction branch, spatial branch, and spatial-guided high-resolution query module. Specifically, a partitioned image patch $I$ and its downscaled version $\hat{I}$ is fed into the encoders for feature extraction. The spatial branch and semantic branch obtain the visual features and semantic features, respectively. In addition, the memory-based interaction branch relies on an external memory bank to mitigate the bias in the guidance information, which associates the cross-image compact semantic representation with high-resolution spatial information for regularization. Inspired by \cite{chen2021learning}, the spatial-guided high-resolution query module is dedicated to infer the high-resolution semantics in a pixel-by-pixel manner by querying the corresponding latent embeddings with the guidance of high-resolution structural information and low-resolution semantics. In the following, we will elaborate on the technical details of the two components.



\subsection{Spatial-guided High-resolution Query Module}

High-quality local image segmentation relies on contextual cues beyond local patches. However, incorporating such context can increase computational overhead. To balance segmentation quality and computational efficiency, we propose a novel spatial-guided high-resolution query module that queries the nearest latent embedding of a given spatial coordinate to obtain the corresponding semantic result without the need for additional context.

In concrete, we first define a query function $f_\theta$ ($\theta$ is learnable parameters) over the feature maps to achieve the high-resolution semantic mask $S$ ($S \in \mathbb{R}^{C \times H \times W}$) where $C$ denotes the number of semantic classes. Here, the feature maps are viewed as latent embeddings evenly distributed in spatial dimensions and we assign corresponding coordinates to them. Hence, the value at the coordinate $x_q$ of the high-resolution semantic mask $S$ can be queried as below:
\begin{align}
    S(x_q) = f_\theta(z^*, x_q-x^*),\label{eq:mask}
\end{align}
where $z^*$ is the nearest latent embedding from $x_q$ and $x^*$ is the low-resolution coordinate of the latent embedding $z^*$ in the spatial domain. Given the relative coordinates, the high-resolution query function $f_\theta$ can query the nearest semantic result set of the latent embedding $z^*$.

To this end, we adopt a vanilla MLP as the query function shared by each latent embedding. However, the previous study \cite{xu2021ultrasr} implies that neural networks are insensitive to high-frequency signals and are inclined to learn low-frequency signals. This may lead to undesirable artifacts for intricate ultra-high resolution images. Consequently, we encode the relative coordinates via a periodic function $\phi$ to enhance the capability of the network in the high-frequency domain. Thus, the encoding function of the relative coordinates (i.e., $x_q-x^*$) is defined as:
\begin{align}
    \phi(x_q-x^*) = [\omega_1sin(x_q-x^*),\omega_1cos(x_q-x^*),..., \nonumber \\
    \omega_nsin(x_q-x^*),\omega_ncos(x_q-x^*)],
\end{align}
where $\omega_1,...,\omega_n$ are initially set to $2e^i$ ($i \in [1,...,n]$) as the frequency parameters that will be fine-tuned on the training stage. As such, $\phi(\cdot)$ maps the relative coordinate to a $2n$-dimensional positional information. 
Eq.~\ref{eq:mask} can be rewritten as:
\begin{align}
    S(x_q) = f_\theta(z^*, x_q-x^*, \phi(x_q-x^*)).
\end{align}

In general, the spatial resolution of the latent embedding is much smaller than that of a semantic map, which results in the loss of spatial details during the query. To this end, we propose to utilize the higher resolution masks to provide spatial guidance for the latent embeddings. Specifically, the high-resolution segmentation masks $M_b$ ($M_b \in \mathbb{R}^{C \times \frac{H}{2} \times \frac{W}{2}}$) and $M_l$ ($M_l \in \mathbb{R}^{C \times H \times W}$) are computed by the memory-based interaction branch and spatial branch, respectively, which modifies the high-resolution query function as below:
\begin{align}
    S(x_q) = f_\theta(z^*, x_q-x^*, \phi(x_q-x^*), \nonumber \\
    m_b^*, x_q-x_b^*, \phi(x_q-x_b^*),  \nonumber \\
    m_l^*, x_q-x_l^*, \phi(x_q-x_l^*)),
\end{align}
where $m_b^*$ and $m_l^*$ denote the nearest mask values from $x_q$ in mask $M_b$ and $M_l$. $x_b^*$ and $x_l^*$ are the corresponding low-resolution coordinates of mask values $m_b^*$ and $m_l^*$, separately. With the cues of high-resolution spatial information, the latent embeddings can better predict fine semantic results.

\subsection{Memory-based Interaction Scheme}

In our model, the high-resolution spatial mask $M_b$ is computed by the last layer of the memory-based interaction branch. Since this branch is relatively shallow, its estimated high-resolution mask contains semantic bias that may interfere with spatial cues for the query function. 
To mitigate the negative impact, it requires contextual information for regularization. 
Without the need for additional large computation, we propose an efficient memory-based interaction scheme that adds semantics to the mask $M_b$. \wx{It involves an external memory bank $\mathcal{M}$ ($\mathcal{M} \in \mathbb{R}^{D \times C}$) that stores the semantic features across images, where $D$ denotes the feature dimension.}
To reduce the computation overhead to the largest extent, our proposed memory scheme costs linear complexity only. 

As the preliminary step, we randomly select an image sample from the training set. Then, we calculate a mean vector of the semantic branch features for each category as the initial value of the memory bank, with the aid of the ground-truth segmentation maps. During training, the representation of each category $c$ ($c \in [1,...,C]$) in the memory bank $\mathcal{M}$ is updated by the moving average method in the $t$-th iteration.
\begin{align}
    \mathcal{M}^c_t = m \cdot \mathcal{M}^c_{t-1} + (1-m) \cdot \varphi(\mathcal{R}_{t-1}),
\end{align}
where the momentum $m$ is set as 0.9, and $\varphi$ is a transform function. $\mathcal{R}_t$ ($\mathcal{R}_t \in \mathbb{R}^{D \times \frac{H}{4} \times \frac{W}{4}}$) is the semantic branch features of the current sample in the $t$-th iteration. In $\varphi$, $\mathcal{R}_{t-1}$ is permuted with the dimension $D \times N$ ($N=\frac{HW}{16}$). 

The feature representation of each category in the memory bank can be denoted as 
$\mathcal{R}^c$ ($\mathcal{R}^c \in \mathbb{R}^{D \times N^c}$) that stores the representation of the category $c$, where $N^c$ is the number of the pixels labeled as the category $c$. $\mathcal{GT}$ ($\mathcal{GT} \in \mathbb{R}^{D \times N}$) stores the ground-truth category labels of $\mathcal{R}$. Next, we calculate the cosine similarity $\mathcal{S}^c$ ($\mathcal{S}^c \in \mathbb{R}^{N^c}$) between $\mathcal{R}^c$ and $\mathcal{M}^c$:
\begin{align}
    \mathcal{S}^c = \frac{\mathcal{R}^c \cdot \mathcal{M}^c}{\|\mathcal{R}^c\|_2 \cdot \|\mathcal{M}^c\|_2}.
\end{align}
Finally, the transform function $\varphi$ outputs:
\begin{align}
    \hat{\mathcal{R}}^c = \sum^{N^c}_{i=1}\frac{1-\mathcal{S}^c_i}{\sum^{N^c}_{j=1}(1-\mathcal{S}^c_j)} \cdot \mathcal{R}^c_i.
\end{align}

After updating the memory bank, we associate the memory bank as the compact cross-image semantic representation to high-resolution information to enhance the mask $M_b$ in the semantic perspective. Specifically, we read the memory bank $\mathcal{M}$ and extract the features $\mathcal{F}_b$ ($\mathcal{F}_b \in \mathbb{R}^{D \times \frac{H}{2} \times \frac{W}{2}}$) from the memory-based interaction branch. Then, $\mathcal{F}_b$ is permuted into the features with the dimension $D \times \frac{HW}{4}$. Thus, we calculate the relation $\mathcal{W}$:
\begin{align}
    \mathcal{W} = \text{Softmax}(\frac{\mathcal{M}^\top \otimes \mathcal{F}_b}{\sqrt{D}}),
\end{align}
where $\otimes$ is matrix multiplication. The size of $\mathcal{W}$ is $C \times \frac{HW}{4}$ and it is reshaped as $C \times \frac{H}{2} \times \frac{W}{2}$. Last, $M_b$ is refined as follows:
\begin{align}
    M_b = (1+\mathcal{W}) \cdot M_b.
\end{align}
\qi{Note that, compared with the previous memory scheme with quadratic complexity (e.g. \cite{jin2022mcibi++}), our proposed memory scheme has a linear computational complexity of $\mathcal{O}(\frac{HWC}{4})$. This is because our scheme does not associate the global pixel memory for each pixel, which spares the self-attention computation, contributing significantly to a substantial increase in overall computational efficiency.} 

\section{Experimental Results}

\begin{figure*}[t]
    \centering
    
    \begin{tabular}{c@{}c}
        \includegraphics[width=0.48\textwidth]{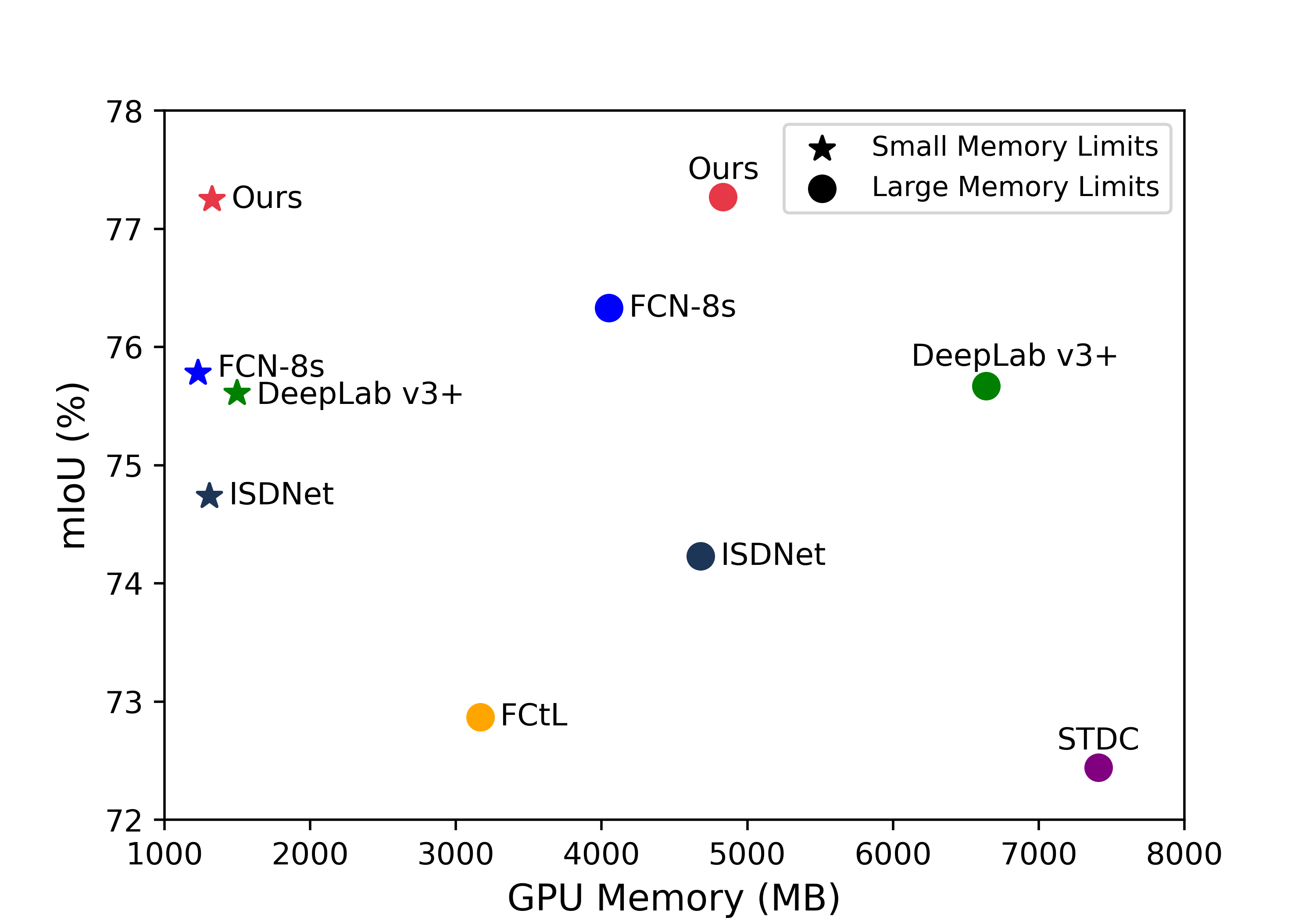} &
        \includegraphics[width=0.48\textwidth]{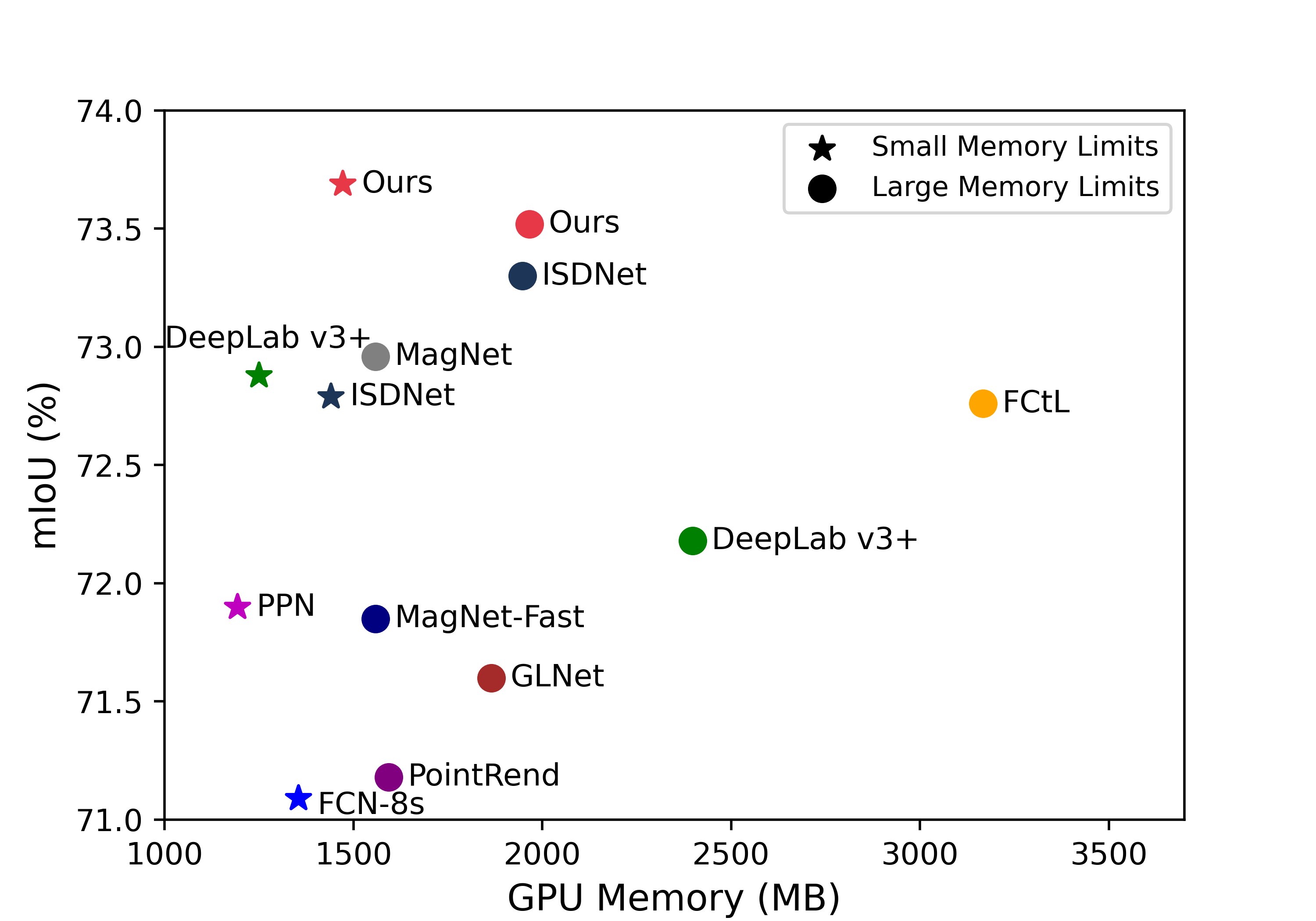} \\
        {\footnotesize Comparison on Inria Aerial} & {\footnotesize Comparison on DeepGlobe}
    \end{tabular}
    \vspace{-0.1cm}
    \caption{Comparison of mIoU v.s. GPU Memory cost under small and large GPU memory constraints (denoted as stars and circles). }
    \label{fig:all}
\end{figure*}

In this section, we conduct experiments over two public benchmarks to evaluate our problem and the capability of the proposed model. We first compare our model against the previous state-of-the-art methods and then perform ablation studies to delve into our model structure.

\subsection{Datasets}

\textbf{Inria Aerial \cite{maggiori2017can}.} This dataset contains large resolution aerial images of five cities, ranging from dense metropolitan districts to alpine resorts. It contains 180 aerial images of 5000$\times$5000 pixels with the binary mask for building/non-building areas. Following the protocol of \cite{chen2019collaborative}, we split images into training, validation, and testing sets with 126, 27, and 27 images, respectively.

\textbf{DeepGlobe \cite{demir2018deepglobe}.} This dataset provides 803 ultra-high resolution aerial images with 2448$\times$2448 pixels. It contains seven classes of landscape regions, including urban, agriculture, rangeland, forest, water, barren, and unknown region not considered in the challenge. We split all images following \cite{chen2019collaborative}, i.e., 454 training images, 207 validation images, and 142 testing images.


\subsection{Implementation Details}


\textbf{Training details.} We implement our framework using the mmSegmentation \cite{contributors2020mmsegmentation} toolbox on a workstation with a single NVIDIA RTX 3090 GPU. In particular, we adopt DeepLabv3 \cite{chen2017rethinking} with ResNet18 \cite{he2016deep} as the encoder of the semantic branch and STDC \cite{fan2021rethinking} as the encoder of the memory-based interaction branch. During training, we optimize the parameters adapting Stochastic Gradient Descent (SGD) and set the batch size to 4 and 8 for Inria Aerial and DeepGlobe, respectively. The initial learning rate is set to $10^{-2}$, which is decayed by a poly learning rate policy with the power of 0.9. In practice, it takes 40k and 80k iterations to converge our model for two datasets, respectively.

\textbf{Inference details.} During inference, 
we measure the GPU Memory and Frames-per-second (FPS) on an RTX 2080Ti GPU and adopt the same environment as \cite{guo2022isdnet} (i.e., CUDA 10.1, CuDNN 7.6.5, and Pytorch 1.6.0) for fair comparison.

\subsection{Comparison with State-of-the-arts}

In practice, segmenting an ultra-high resolution image often consumes an exceedingly large amount of GPU memory resources. As a consequence, for the robotic platforms like UAVs, they usually have limited computation resources and GPU memory for segmenting and analyzing the large images on board. 
To simulate the challenging situations, in the experiments, we compare the segmentation methods under small and large GPU memory limits.
In this experiment, we set 1.5 GB as the small GPU memory limit. For large GPU memory limits, we set 7.5 GB for Inria Aerial and 3.5 GB for DeepGlobe, due to their different image resolutions. In the following, we conduct the comparison experiments and discuss the results.

\textbf{Small memory limits.} In general, there are two ways to segment the ultra-high resolution images: 1) segmenting downsampled global images (denoted as \textit{global inference}); and 2) cropping, segmenting, and merging local patches (denoted as \textit{local inference}).
We first consider the methods that adopt the global inference strategy, which straightforwardly downsamples the whole input image $\dot{I}$ to meet the requirements of small limited GPU memory. To this end, we retrain and test U-Net \cite{ronneberger2015u}, FCN-8s \cite{long2015fully}, and DeepLab v3+ \cite{chen2018encoder} on Inria Aerial and DeepGlobe datasets. As shown in Table~\ref{tab:com_inriav1} and Table~\ref{tab:com_deepglobev1}, these methods can hardly achieve satisfactory accuracy despite their high FPS, since compressing large images causes severe detail lost. 

\begin{figure*}[t]
    \centering
    \footnotesize
    \vspace{-.5cm}
	\begin{tabular}{c@{}c@{}c@{}c@{}c@{}c}

        \includegraphics[width=0.16\textwidth,height=2.9cm]{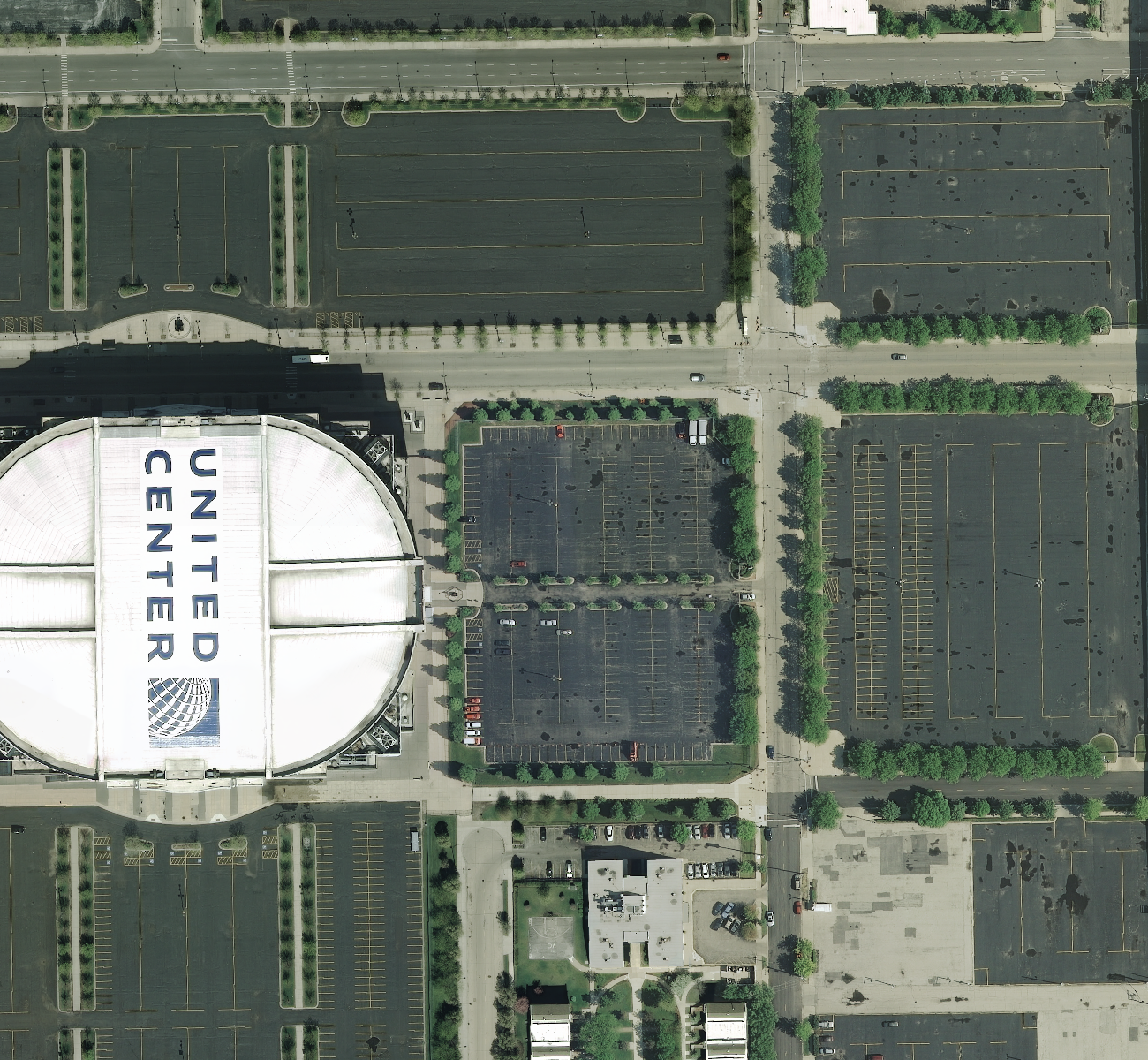} &
	\includegraphics[width=0.16\textwidth,height=2.9cm]{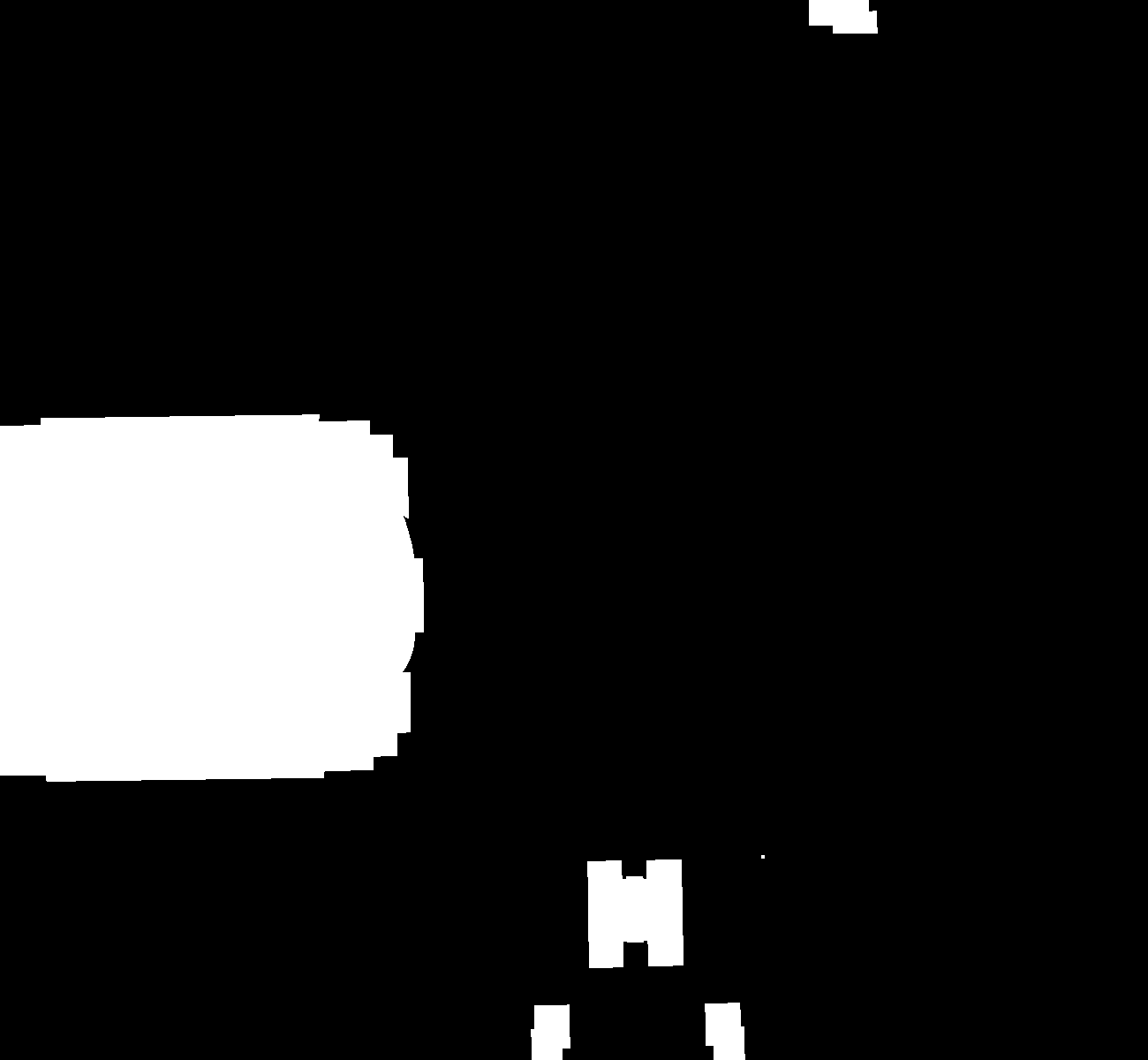} &
        \includegraphics[width=0.16\textwidth,height=2.9cm]{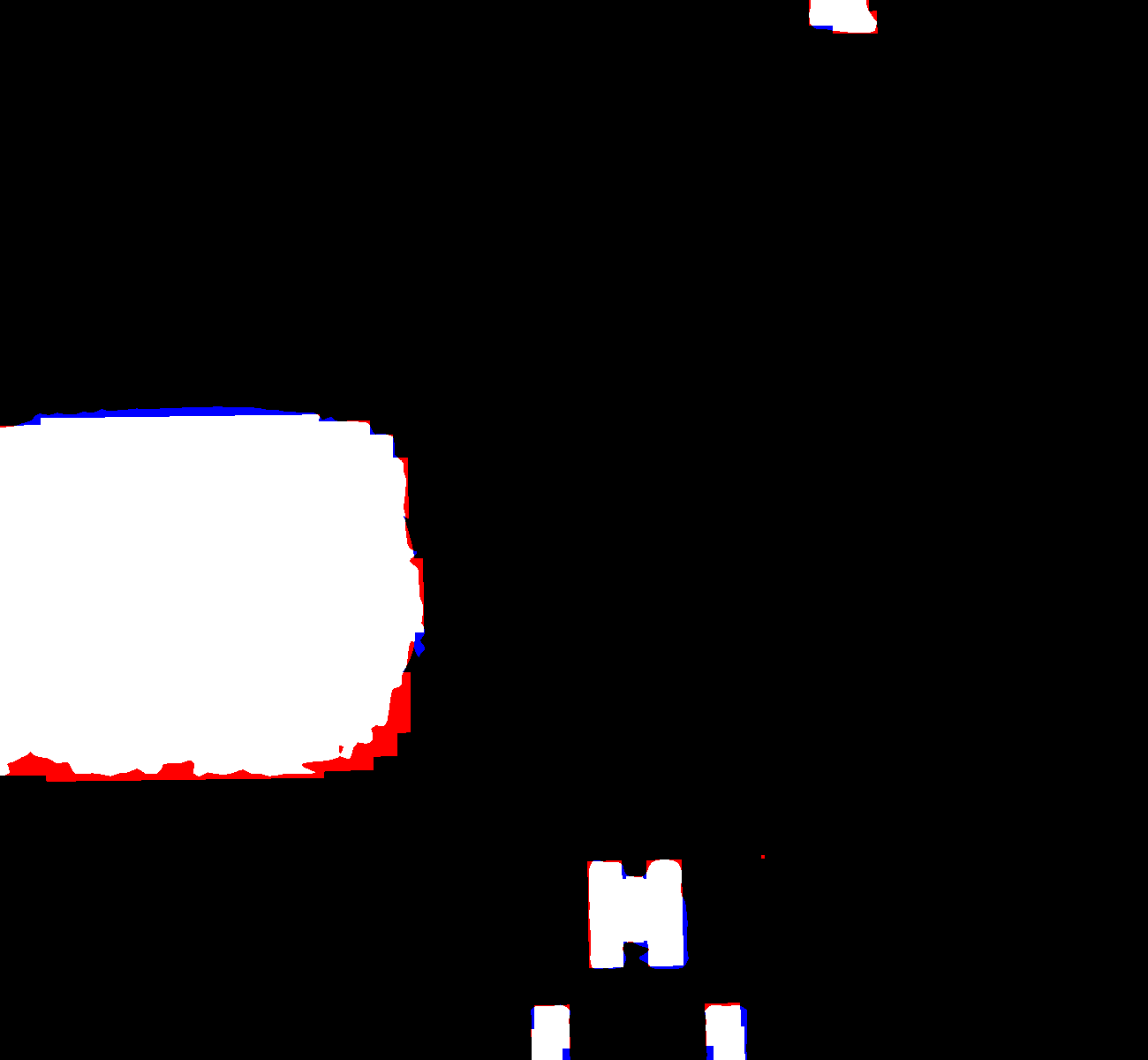} &
        \includegraphics[width=0.16\textwidth,height=2.9cm]{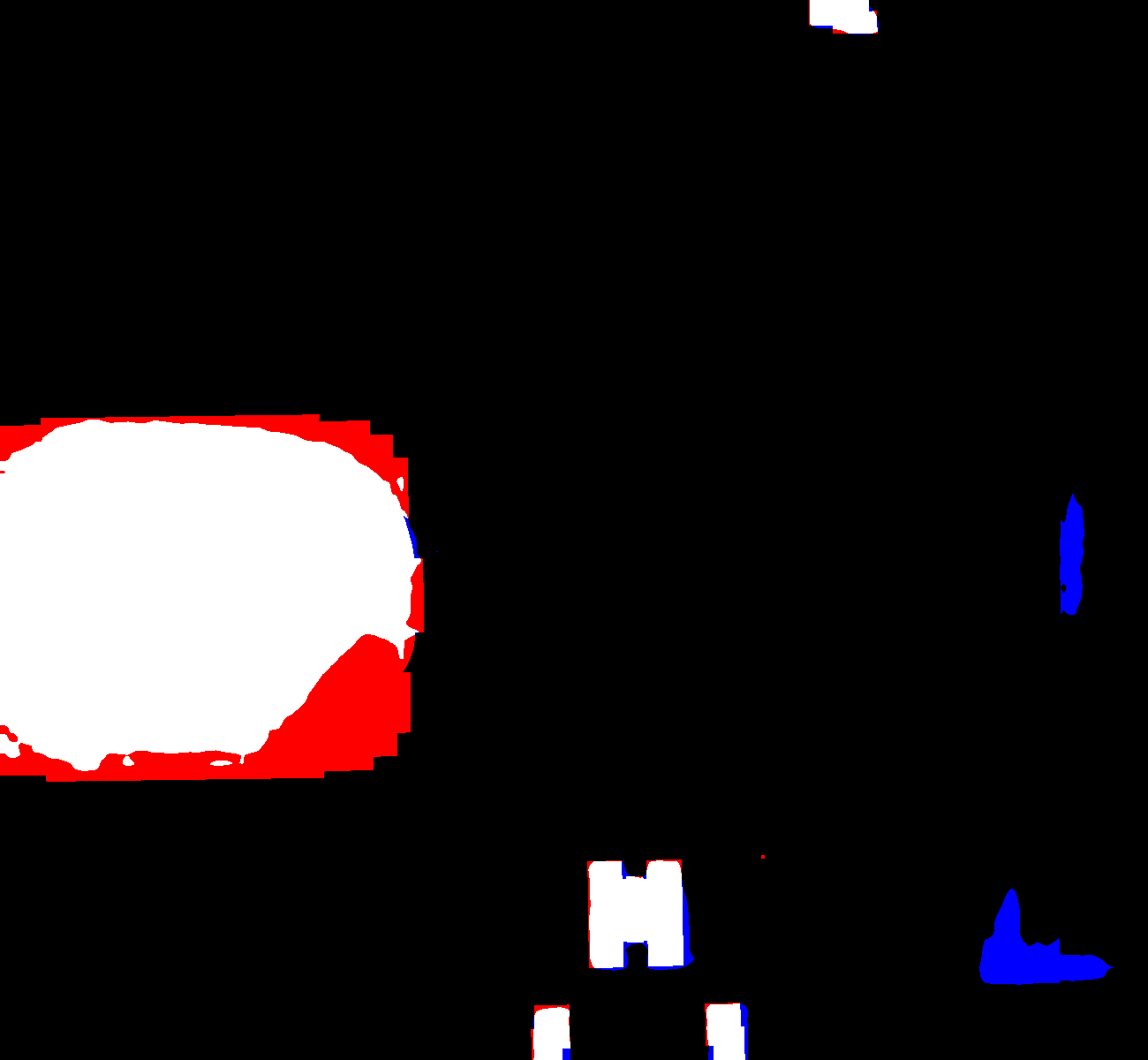} &
        \includegraphics[width=0.16\textwidth,height=2.9cm]{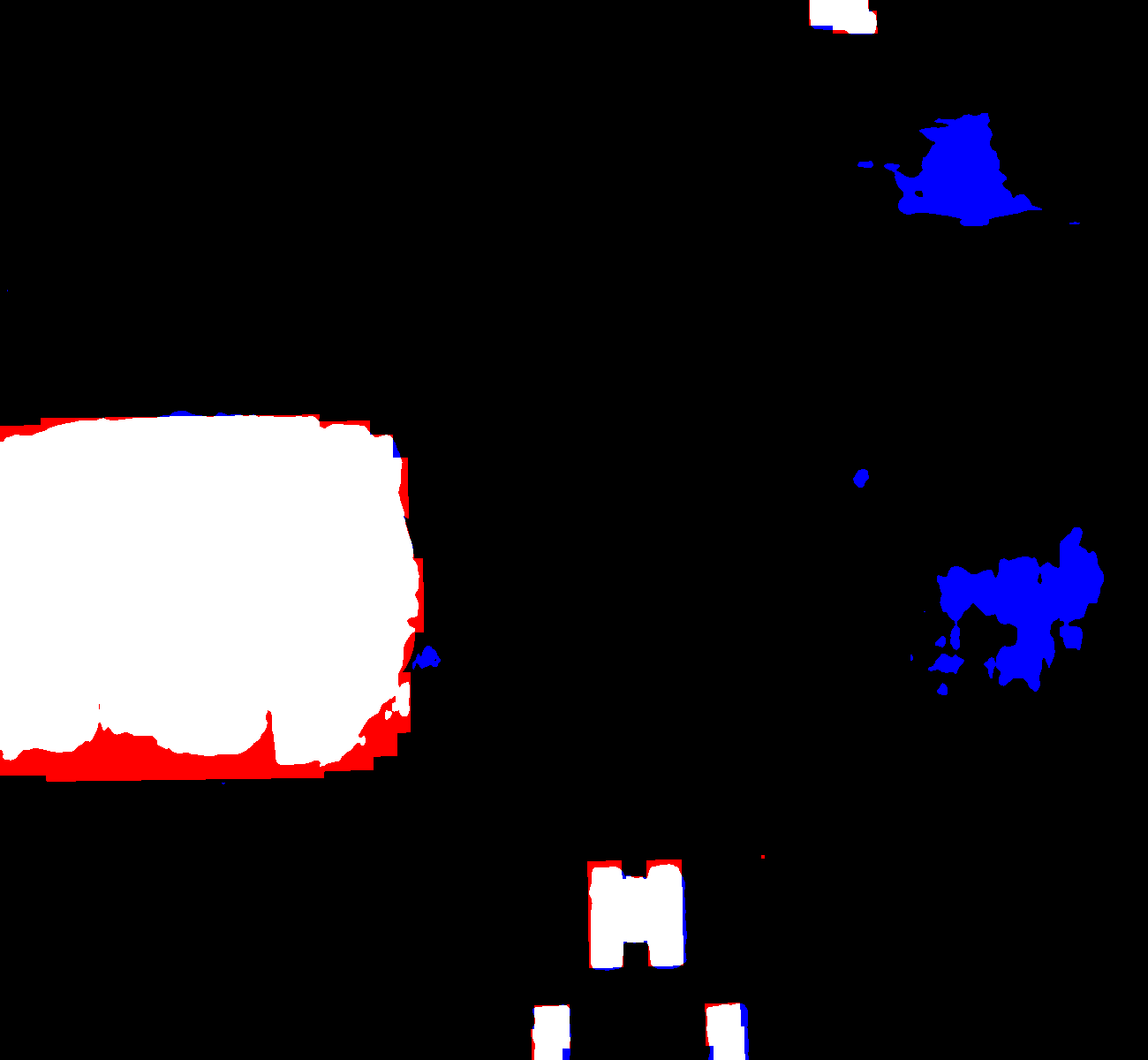} &
        \includegraphics[width=0.16\textwidth,height=2.9cm]{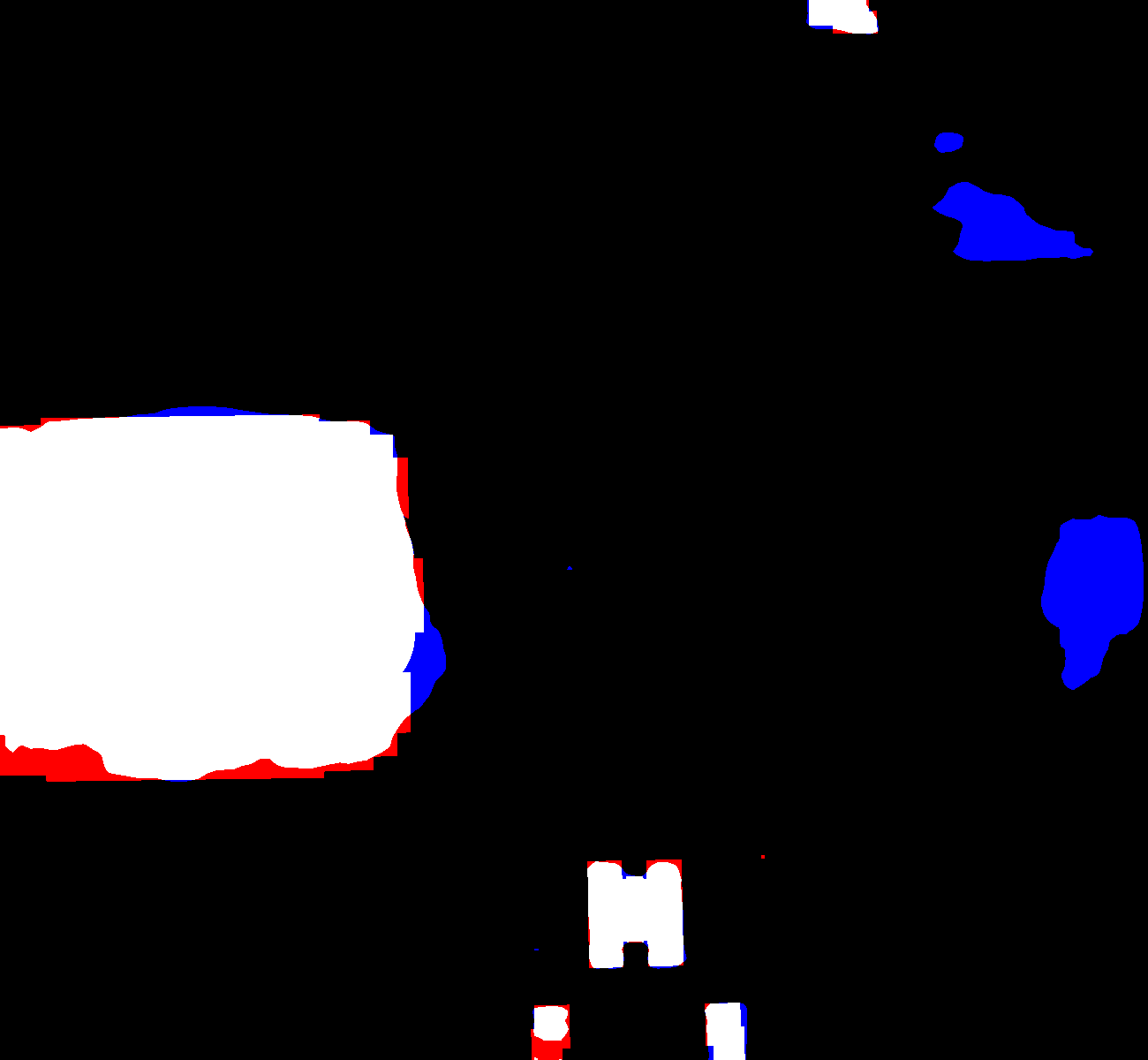} \\

        \includegraphics[width=0.16\textwidth,height=2.9cm]{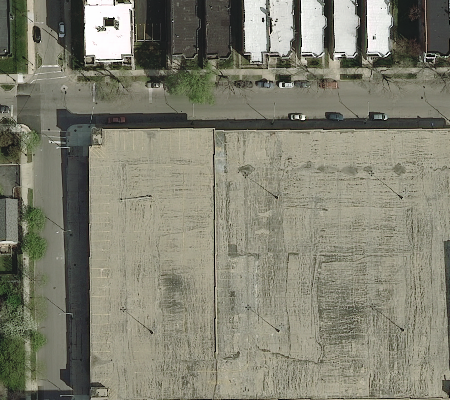} &
	\includegraphics[width=0.16\textwidth,height=2.9cm]{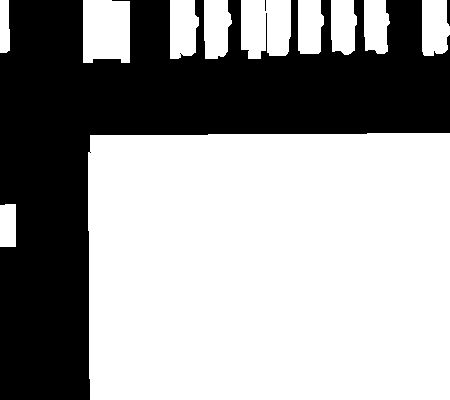} &
        \includegraphics[width=0.16\textwidth,height=2.9cm]{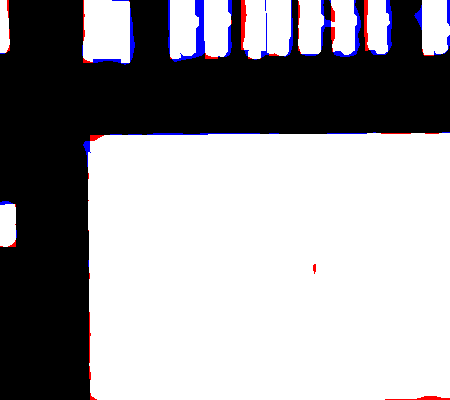} &
        \includegraphics[width=0.16\textwidth,height=2.9cm]{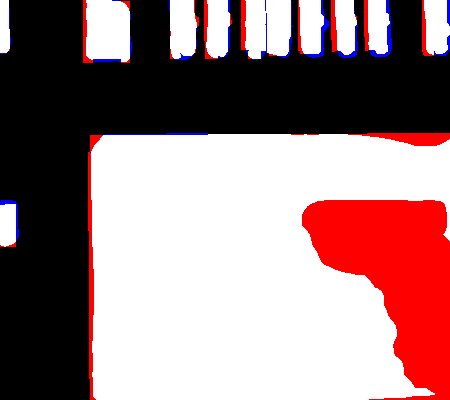} &
        \includegraphics[width=0.16\textwidth,height=2.9cm]{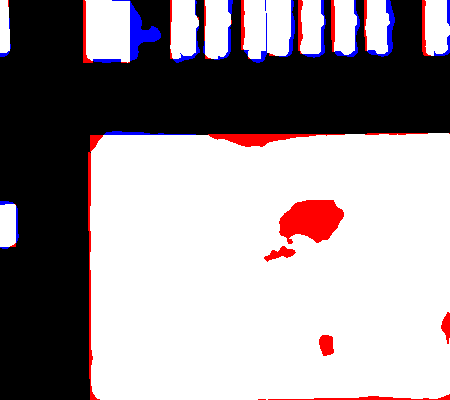} &
        \includegraphics[width=0.16\textwidth,height=2.9cm]{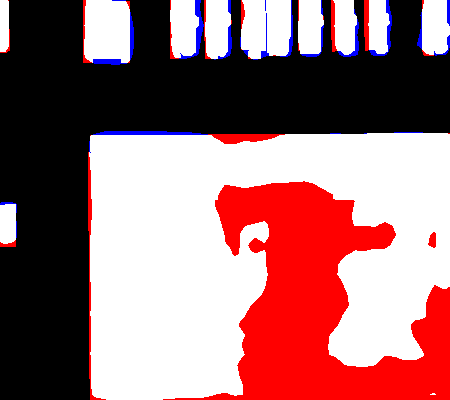} \\

        \includegraphics[width=0.16\textwidth,height=2.9cm]{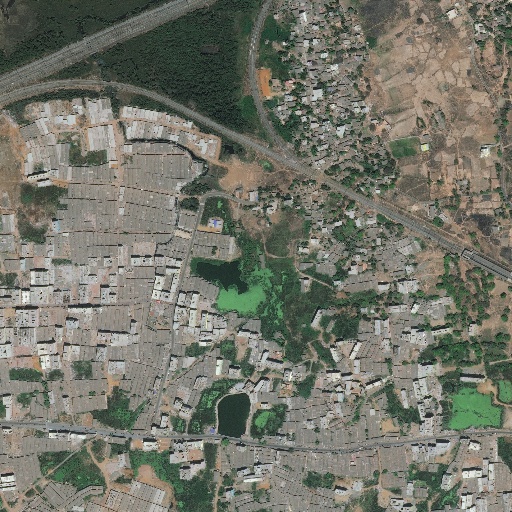} &
	\includegraphics[width=0.16\textwidth,height=2.9cm]{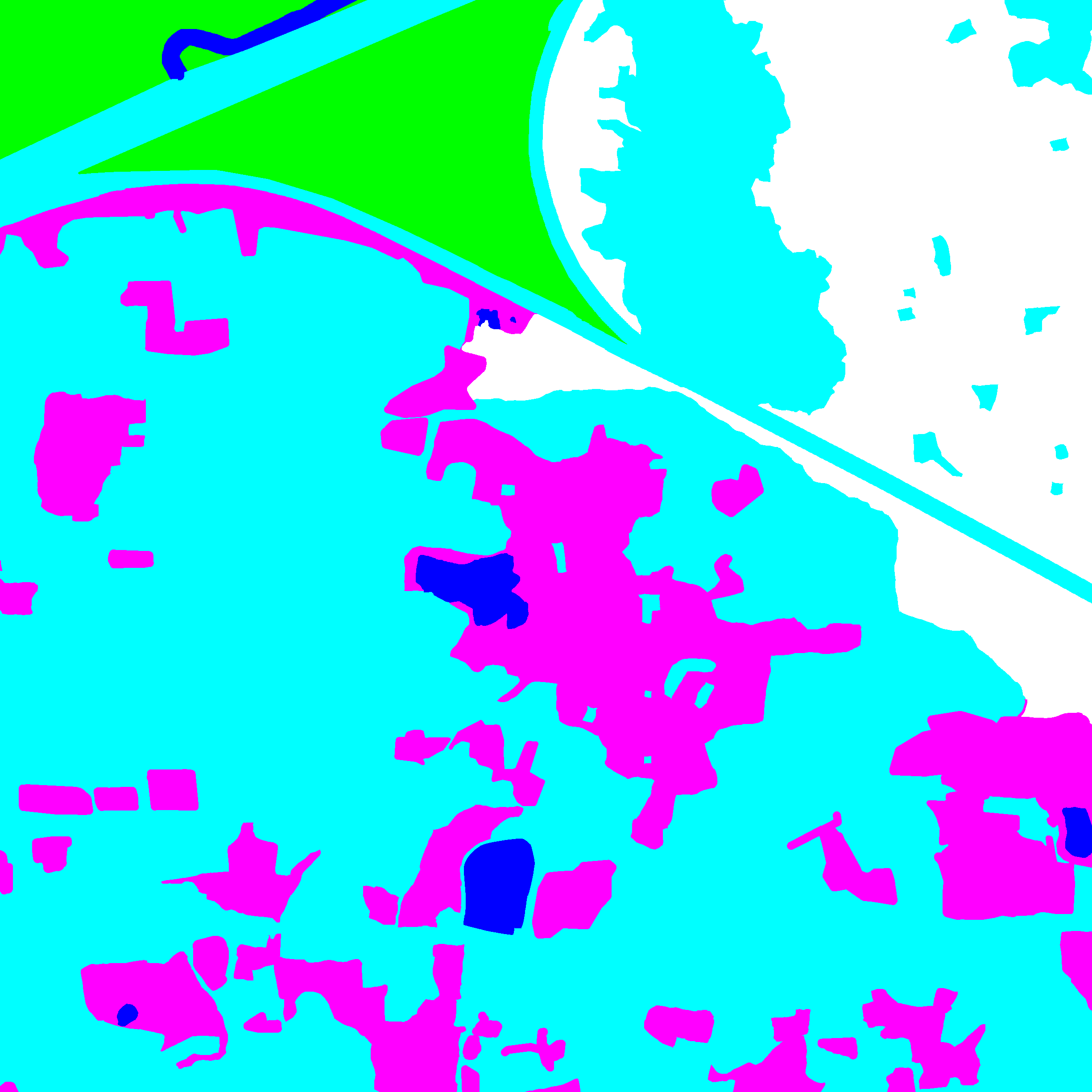} &
        \includegraphics[width=0.16\textwidth,height=2.9cm]{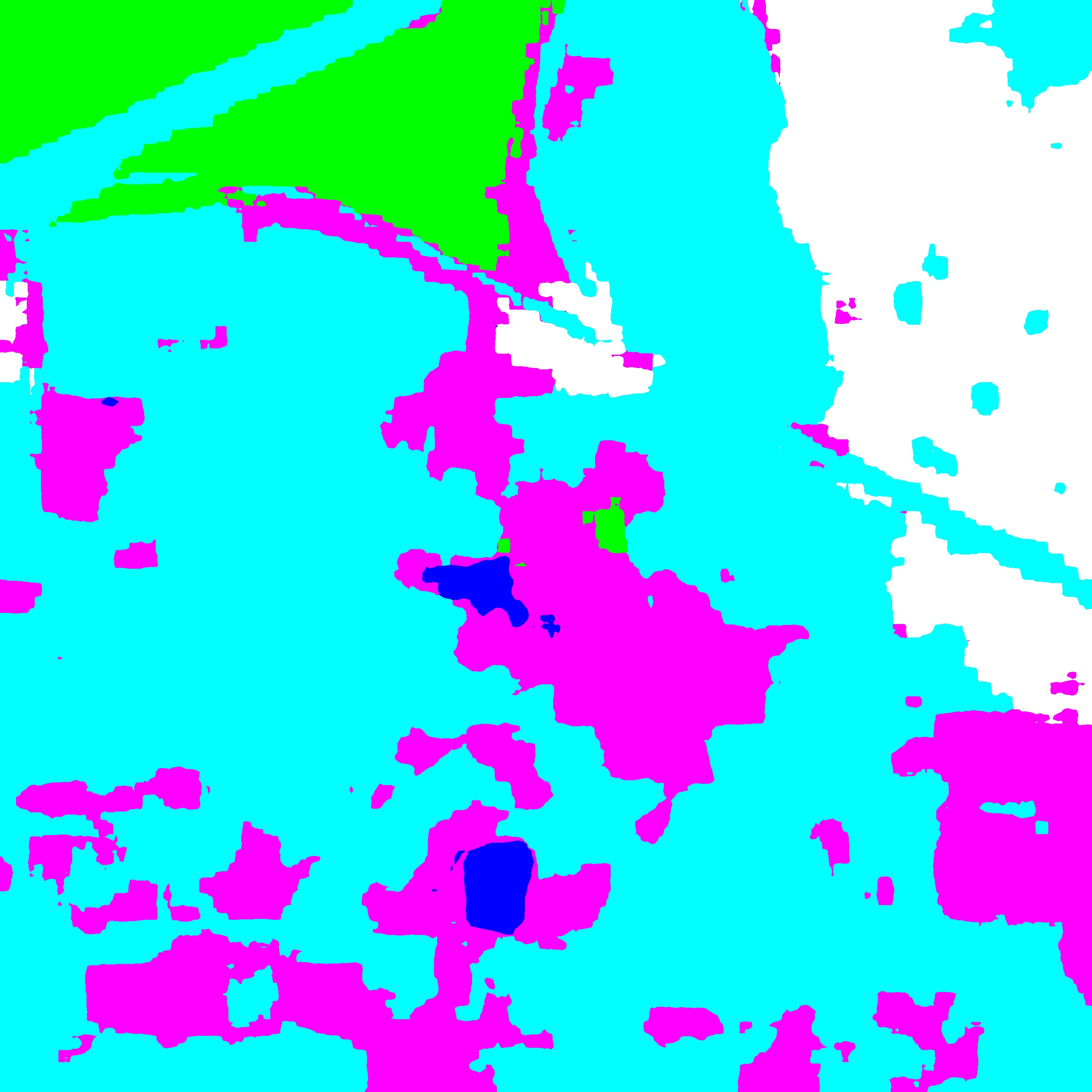} &
        \includegraphics[width=0.16\textwidth,height=2.9cm]{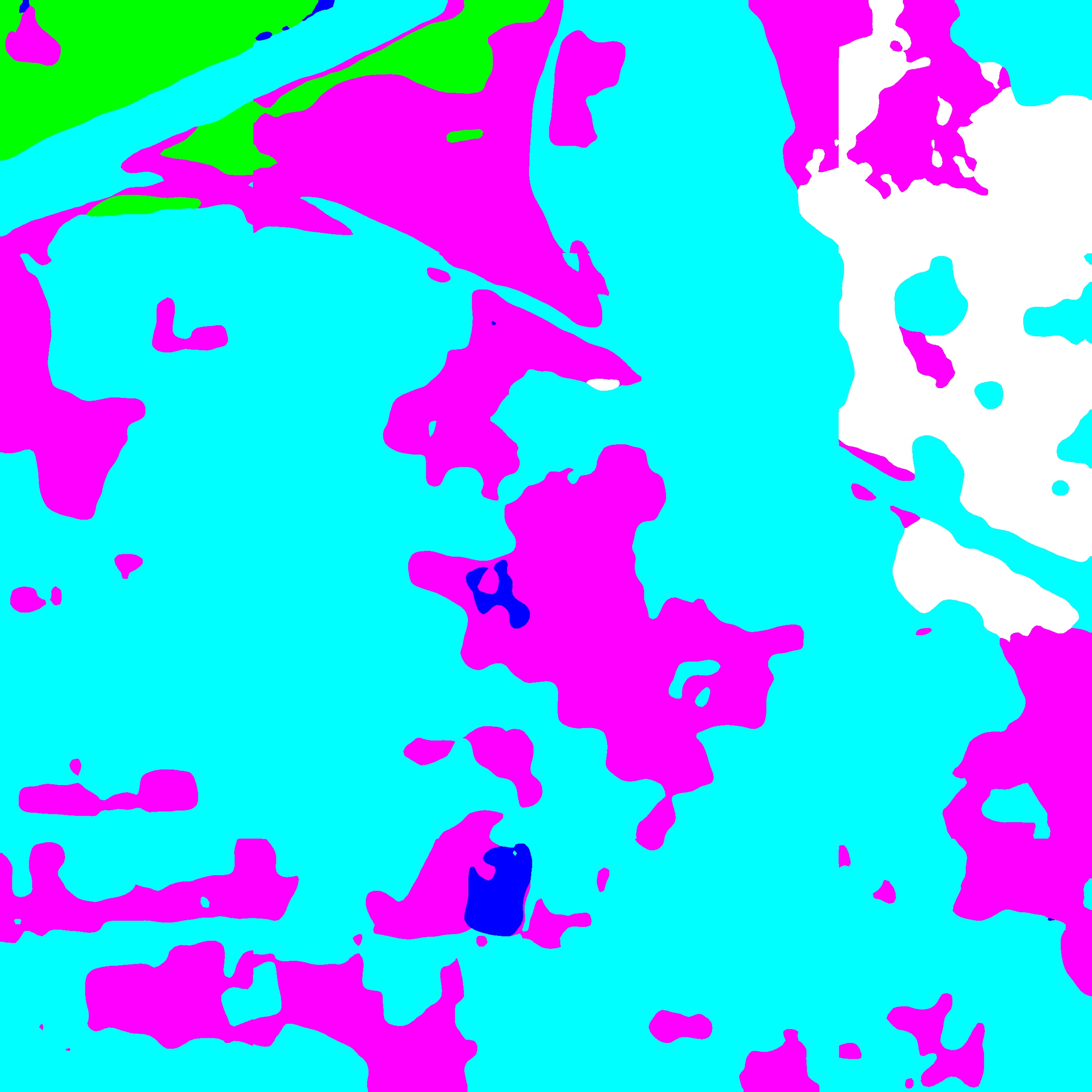} &
        \includegraphics[width=0.16\textwidth,height=2.9cm]{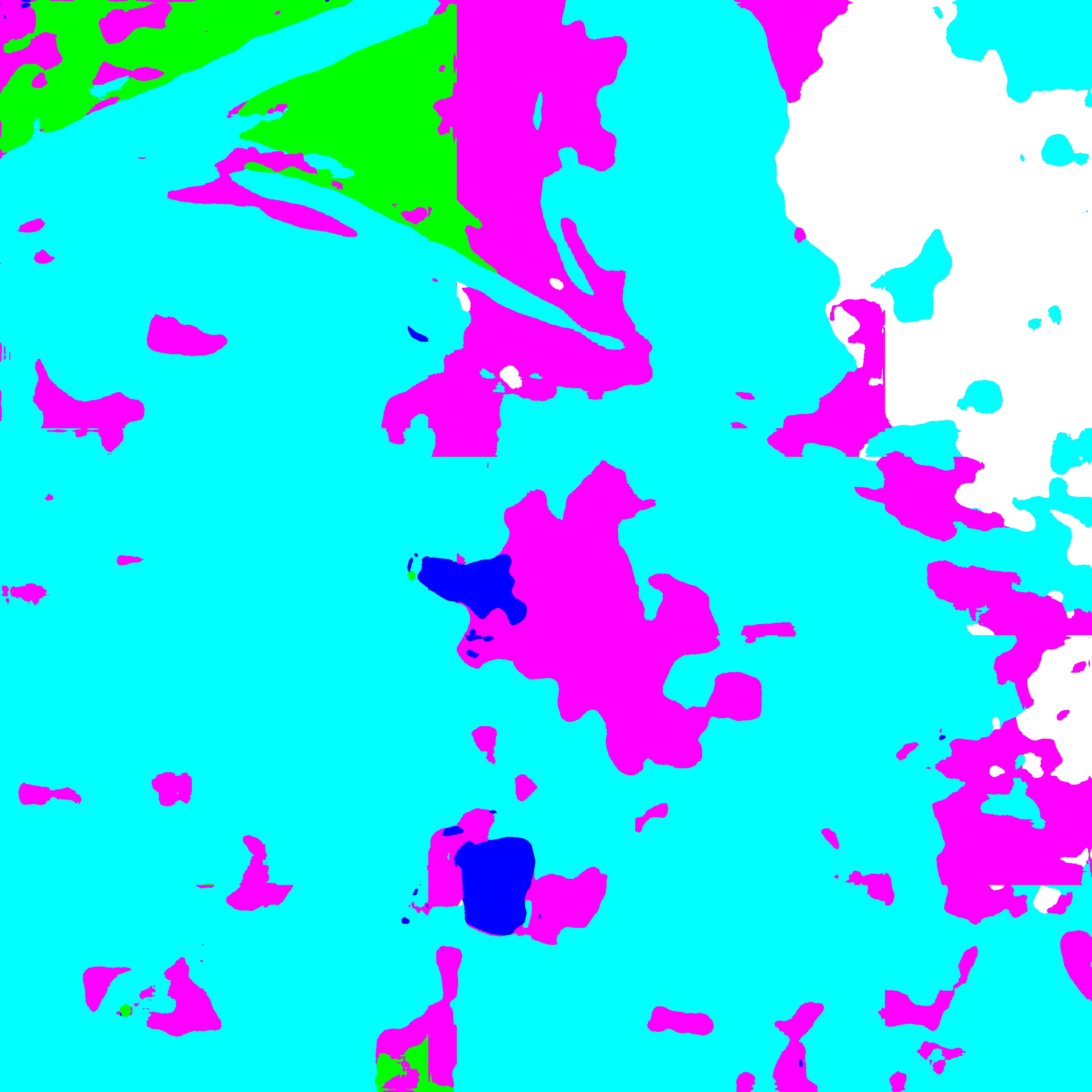} &
        \includegraphics[width=0.16\textwidth,height=2.9cm]{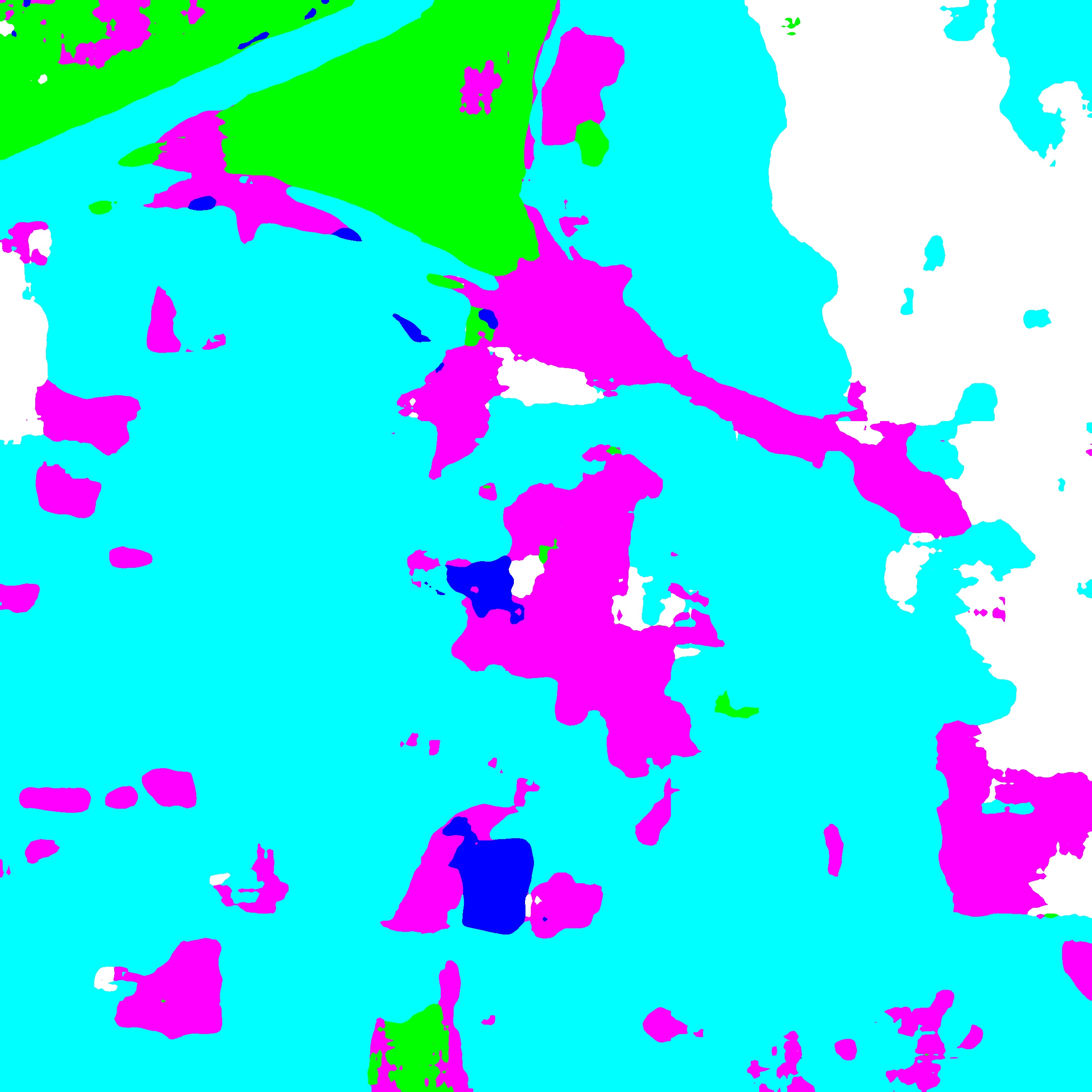} \\

        \includegraphics[width=0.16\textwidth,height=2.9cm]{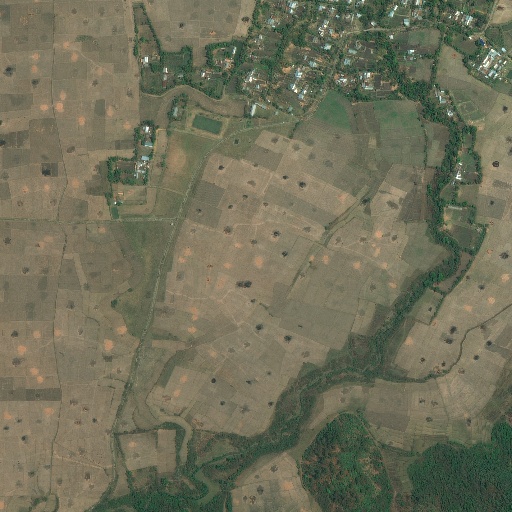} &
	\includegraphics[width=0.16\textwidth,height=2.9cm]{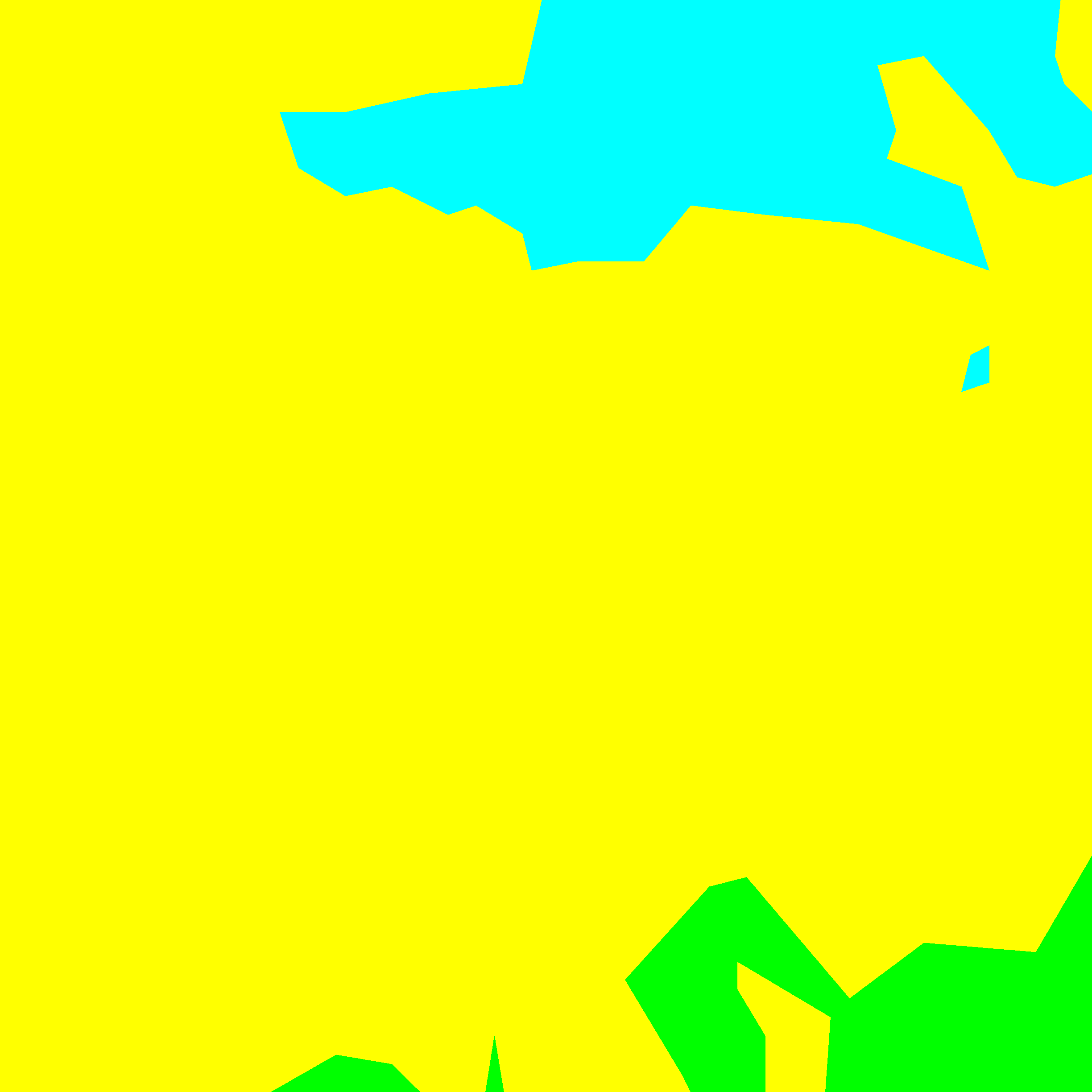} &
        \includegraphics[width=0.16\textwidth,height=2.9cm]{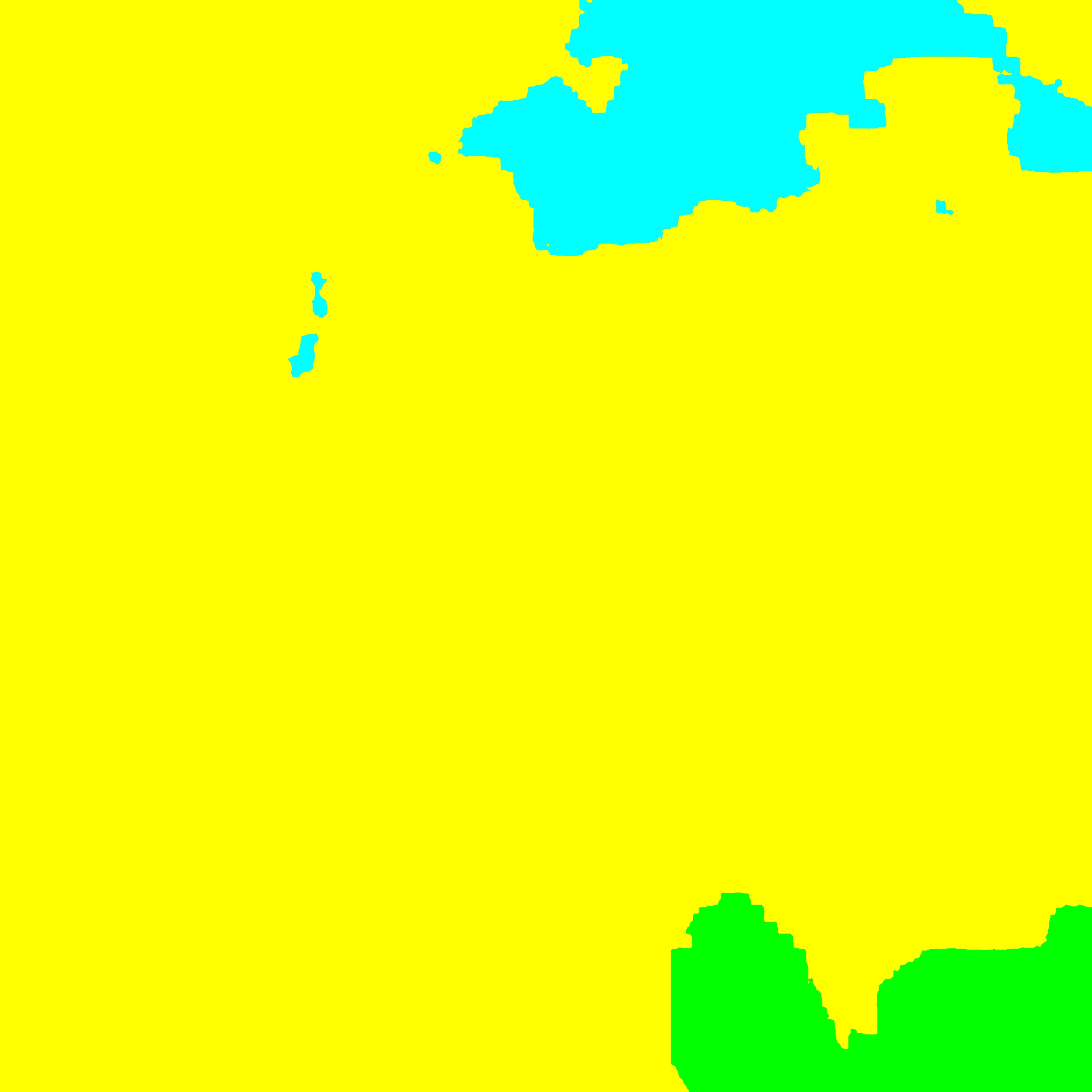} &
        \includegraphics[width=0.16\textwidth,height=2.9cm]{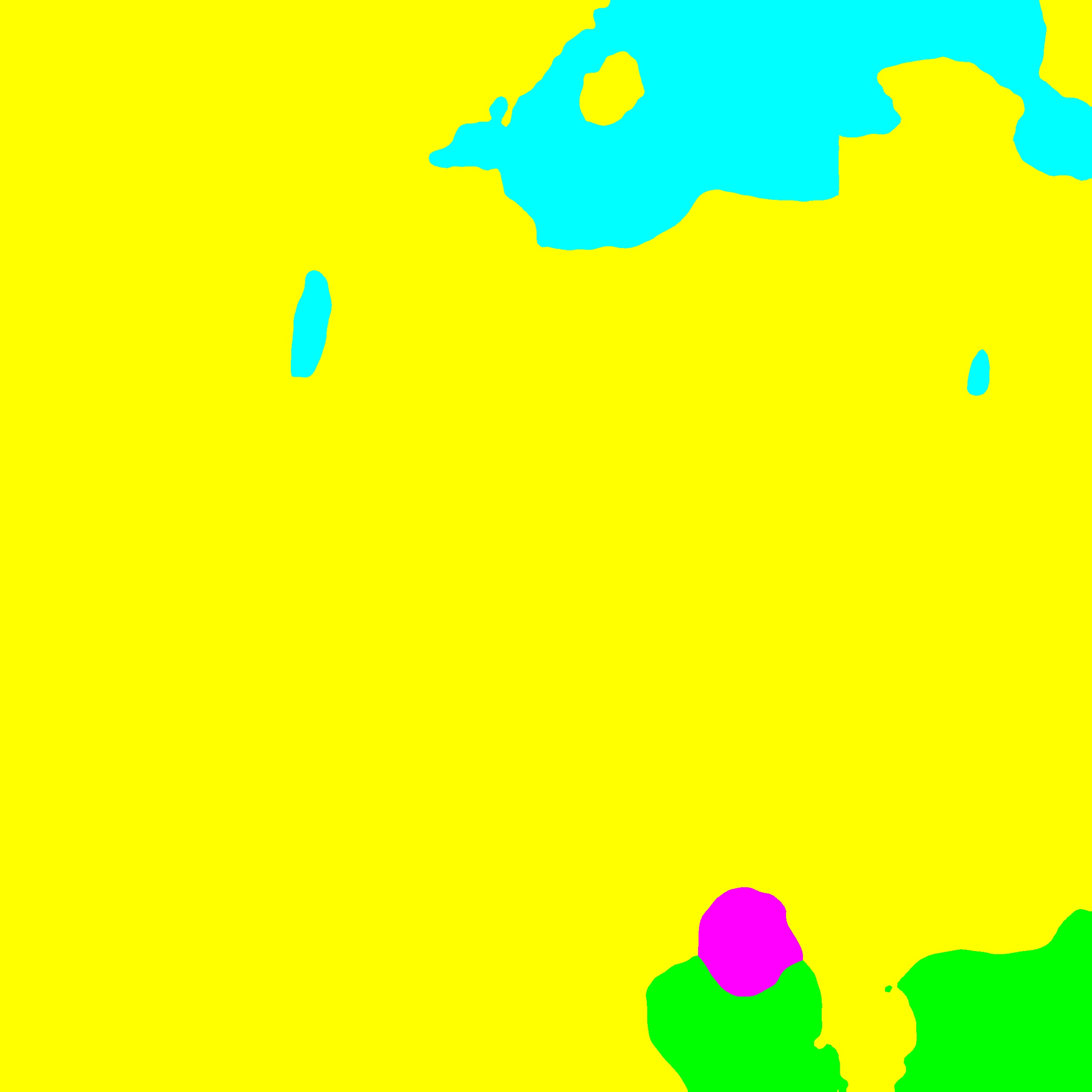} &
        \includegraphics[width=0.16\textwidth,height=2.9cm]{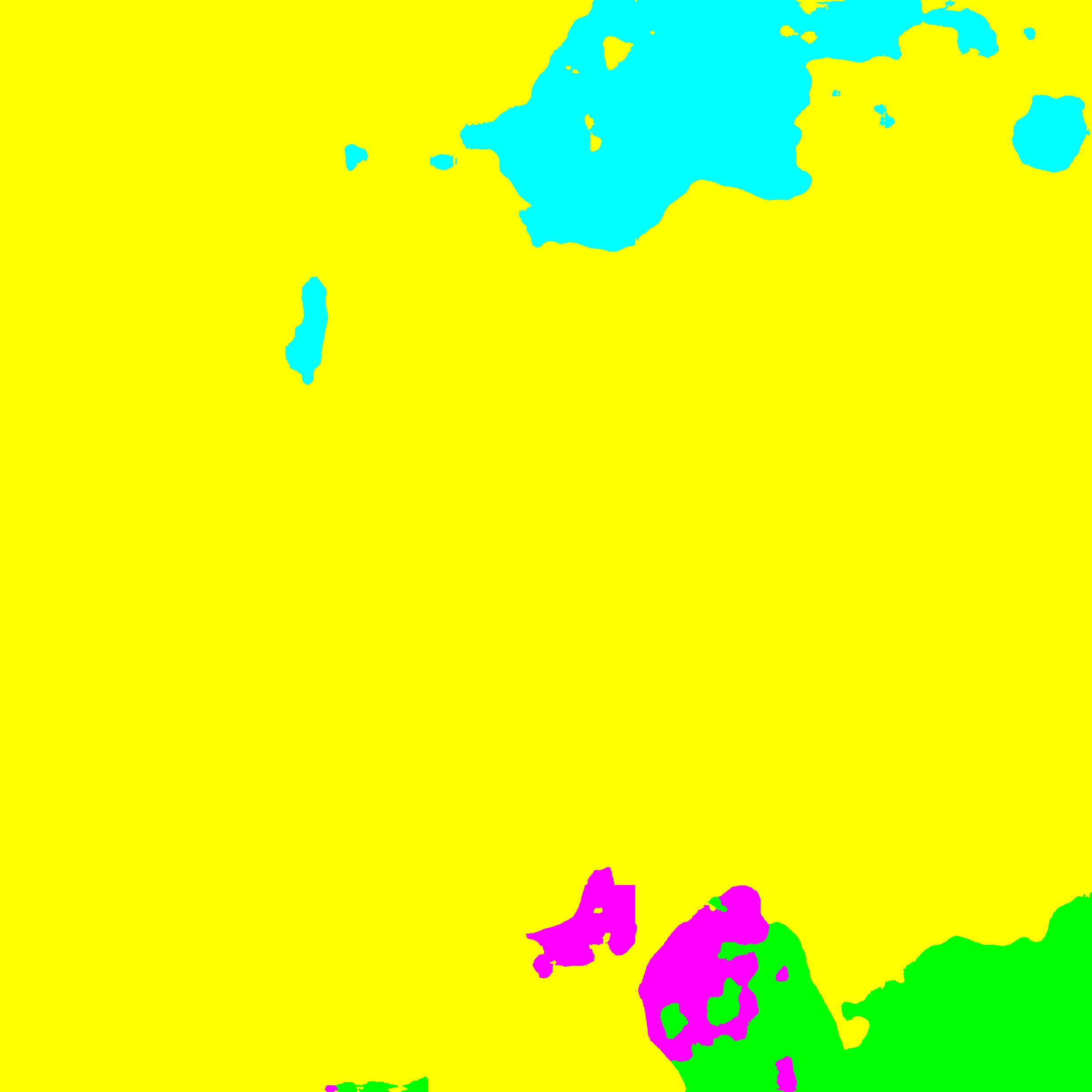} &
        \includegraphics[width=0.16\textwidth,height=2.9cm]{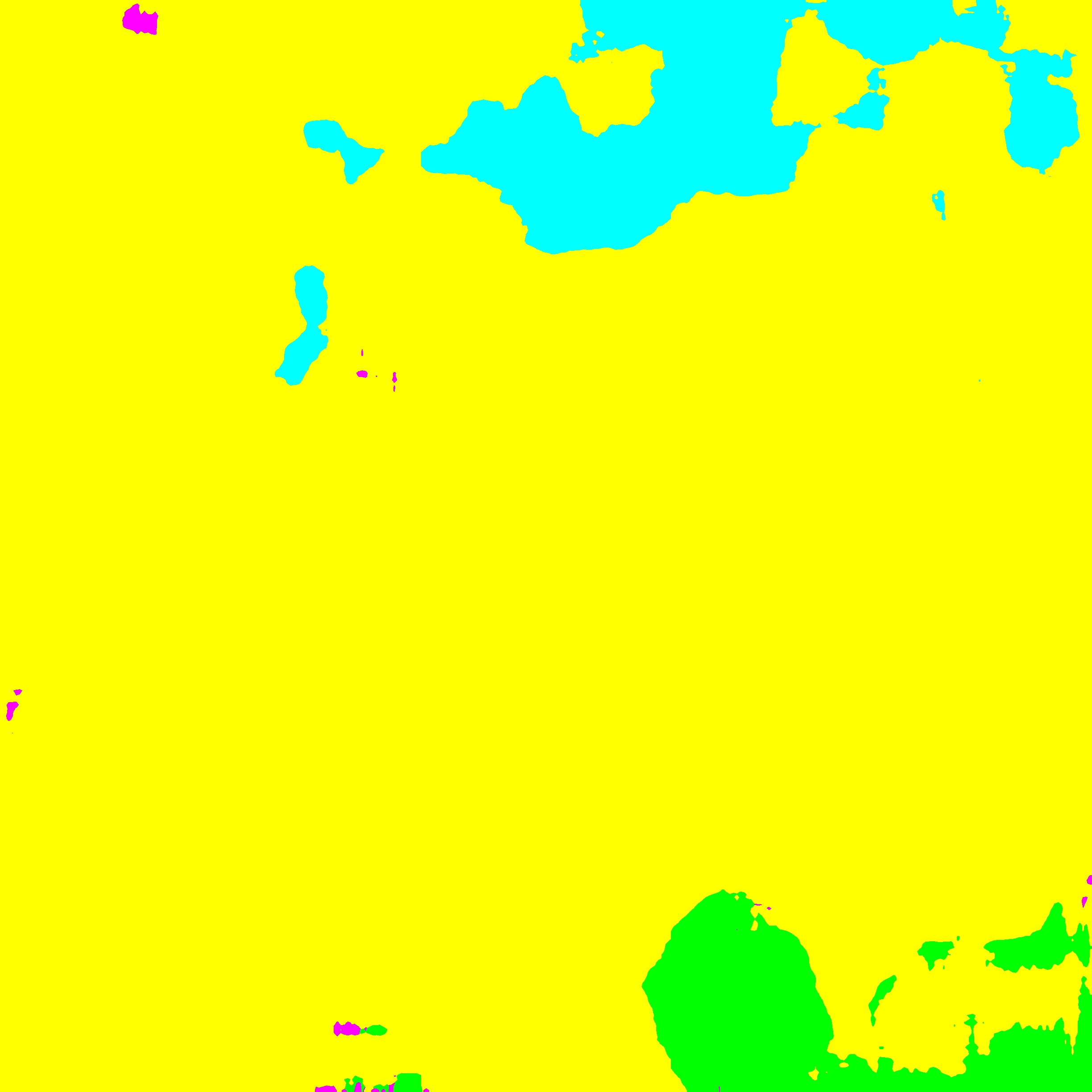} \\
        
        Input Image & Ground-truth & Ours & ISDNet \cite{guo2022isdnet} & Deeplab v3+ \cite{chen2018encoder} & FCN-8s \cite{long2015fully}
        
	\end{tabular}
 \vspace{-0.1cm}
	\caption{
 On the first two rows, we show the representative comparison results from Inria Aerial, where white regions represent the estimated regions for buildings. We highlight the pixels which are the discrepancy between our estimation and ground truth. The blue and red pixels represent False Negatives and False Positives, respectively. On the last two rows, we show representative results from DeepGlobe, where cyan represents "urban", yellow represents "agriculture", purple represents "rangeland", green represents "forest", blue represents "water", and white represents "barren".}
	\label{fig:comparison}
 \vspace{-.6cm}
\end{figure*}

Thus, we adopt the local inference strategy for the segmentation methods to meet small GPU memory limits. 
Table.~\ref{tab:com_inriav1} and Table.~\ref{tab:com_deepglobev1} show our superior performance than other methods in the respective datasets. In specific, our model shows similar GPU memory usage and running time comparing to the state-of-the-art ISDNet \cite{guo2022isdnet}, but elevates 2.50\% mIoU on Inria Aerial and 0.87\% mIoU on DeepGlobe. It is worth noting that our model achieves a significant advantage on Inria Aerial dataset, because our proposed spatial-guided high-resolution query module can depict delicate objects and this dataset contains a large number of small buildings. Besides, as a ResNet-50 based model, PPN \cite{wu2020patch} is also able to fulfill the requirement of limiting small GPU memory, but it underperforms our method and ISDNet. In Fig.~\ref{fig:comparison}, we illustrate several qualitative comparison results. We observe that our model has the ability to delineate fine regions (e.g., small urban and rivers) and stronger semantic discrimination (e.g., large forests and rangeland). This is attributed to the high-resolution query and memory-based semantic enhancement.

\textbf{Large memory limits.} To evaluate the capability of our model under large GPU memory limits, we compare our model with BiSeNetV1 \cite{yu2018bisenet}, STDC \cite{fan2021rethinking}, PointRend \cite{kirillov2020pointrend}, CascadePSP \cite{cheng2020cascadepsp}, GLNet \cite{chen2019collaborative}, MagNet \cite{huynh2021progressive}, FCtL \cite{li2021contexts}, and ISDNet \cite{guo2022isdnet},
in terms of mIoU, GPU Memory, and FPS. The comparison results are depicted in Table.~\ref{tab:com_inriav2} and Table.~\ref{tab:com_deepglobev2}, in which we follow most of the results reported by \cite{guo2022isdnet}. As observed, our model is comparable to the state-of-the-art ISDNet on DeepGlobe, but we gain at least 3\% improvement than it on Inria Aerial. This shows that our approach can further trade off performance, GPU memory, and speed. 

\textbf{Performance-memory trade-off.} We quantify and visualize all the state-of-the-art methods based on the small and large GPU memory limits in Fig.~\ref{fig:all}. We can see that our model can achieve optimal mIoU regardless of small or large GPU memory limits. More importantly, our model can obtain similar performances under small and large GPU memory limits (77.25\% v.s. 77.27\% on Inria Aerial and 73.66\% v.s. 73.50\% on DeepGlobe ), indicating that our robustness to GPU memory constraints, whereas other methods are greatly affected due to narrow view of local patches. This still benefits from our query module, which predicts pixel categories only by the corresponding relative coordinates and the nearest latent embeddings.

\subsection{Ablation Study}

In the following, we conduct ablation studies on the two proposed modules. First, we demonstrate the effectiveness of spatial-guided high-resolution information and memory-based interaction scheme in our model. Next, we analyze different strategies for updating and reading the memory bank in our scheme. Last, we investigate the boundary case of GPU memory usage.

\begin{table}
    \centering
    \setlength{\tabcolsep}{4pt}
    \vspace{-0.3cm}
    \caption{Comparison with state-of-the-arts with small GPU memory limits on Inria Aerial. * represents our implementation.}
    \vspace{-0.3cm}
    \footnotesize
    \begin{tabular}{cccccc}
        \toprule      
        Model & Backbone & Inference & mIoU & Memory & FPS \\
        \midrule
        FCN-8s* \cite{long2015fully} & ResNet-18 & Global & 47.74 & 1412 & 12.72 \\
        DeepLab v3+* \cite{chen2018encoder} & ResNet-18 & Global & 34.73 & 1532 & 16.58 \\
        FCN-8s* \cite{long2015fully} & ResNet-18 & Local & 75.58 & 1228 & 1.23 \\
        DeepLab v3+* \cite{chen2018encoder} & ResNet-18 & Local & 76.22 & 1496 & 1.10 \\
        ISDNet* \cite{guo2022isdnet} & ResNet-18 & Local & 74.75 & 1306 & 3.99 \\
        Ours & ResNet-18 & Local & \textbf{77.25} & 1324 & 3.55 \\
        \bottomrule
    \end{tabular}
    \vspace{-0.3cm}
    \label{tab:com_inriav1}
\end{table}

\begin{table}
    \centering
    \setlength{\tabcolsep}{4pt}
    \caption{Comparison with state-of-the-arts with small GPU memory limits on DeepGlobe. * represents our implementation.}
    \vspace{-0.3cm}
    \footnotesize
    \begin{tabular}{cccccc}
        \toprule
        Model & Backbone & Inference & mIoU & Memory & FPS \\
        \midrule
        U-Net* \cite{ronneberger2015u} & U-Net & Global & 20.61 & 1506 & 42.16 \\
        FCN-8s* \cite{long2015fully} & ResNet-18 & Global & 60.41 & 1438 & 12.07 \\
        DeepLab v3+* \cite{chen2018encoder} & ResNet-18 & Global  & 52.43 & 1532 & 16.58 \\ 
        U-Net* \cite{ronneberger2015u} & U-Net & Local & 69.72 & 1426 & 0.65 \\
        FCN-8s* \cite{long2015fully} & ResNet-18 & Local & 71.09 & 1354 & 3.47 \\
        DeepLab v3+* \cite{chen2018encoder} & ResNet-18 & Local & 72.53 & 1250 & 3.06 \\   
        PPN \cite{wu2020patch} & ResNet-50 & Local & 71.90 & 1193 & 12.90 \\
        ISDNet* \cite{guo2022isdnet} & ResNet-18 & Local & 72.79 & 1440 & 11.47 \\
        Ours & ResNet-18 & Local & \textbf{73.66} & 1472 & 10.09 \\
        \bottomrule
    \end{tabular}
    \label{tab:com_deepglobev1}
\end{table}

\textbf{Model structure.} First, we take the naive query module as our baseline, as shown in Table~\ref{tab:ab_inr}, our model just costs 1284MB GPU memory with high FPS, but only gains 69.24\% mIoU. To improve performance, we use STDC \cite{fan2021rethinking} to extract high-resolution spatial information $M_b$. It is able to boost the performance from 69.24\% to 72.06\%, which exceeds bilinear interpolation with 71.81\% mIoU and demonstrates the effectiveness of the spatial mask guidance. Besides, the GPU memory increase is only 186MB and the total GPU memory is still less than 1.5GB. Based on this, we add memory bank $\mathcal{M}$ and higher resolution information $M_l$, separately. It is observed that there is a great performance improvement, especially the guide of spatial information $M_l$ (at least 1.1\% boost). Importantly, the GPU memory and time hardly grow (about 1MB and less than 0.5 FPS) because both are linear operations. Next, with the aid of both, our model can achieve the best result (i.e., 73.66\% mIoU and 1472MB GPU Memory). Besides, we also try to enhance the semantics of higher resolution masks $M_l$ using a new memory bank $\mathcal{M}_l$, but the result gets worse, we think that this mask mainly contains spatial details and the semantics are too weak to improve. Particularly, we can also adopt high-resolution features as the spatial information guide, that is implicit feature alignment (IFA)~\cite{hu2022learning}. Table~\ref{tab:ab_inr} shows IFA is inferior to our method by at least 1.4\% mIoU and uses 270MB more GPU memory than ours. This largely reflects that our method is more effective and efficient. 

\begin{table}
    \centering
    \setlength{\tabcolsep}{4pt}
    \vspace{-0.3cm}
    \caption{Comparison with state-of-the-arts with large GPU memory limits on Inria Aerial. * represents our implementation.}
    \vspace{-0.3cm}
    \footnotesize
    \begin{tabular}{cccccc}
        \toprule      
        Model & Backbone & Inference & mIoU & Memory & FPS \\
        \midrule
        FCN-8s* \cite{long2015fully} & ResNet-18 & Global & 75.67 & 4050 & 1.96 \\
        DeepLab v3+* \cite{chen2018encoder} & ResNet-18 & Global & 76.33 & 6638 & 1.75 \\
        STDC \cite{fan2021rethinking} & STDC & Global & 72.44 & 7410 & 4.97 \\
        CascadePSP \cite{cheng2020cascadepsp} & ResNet-50 & Local & 69.40 & 3236 & 0.03 \\
        GLNet \cite{chen2019collaborative} & ResNet-50 & Local & 71.20 & 2663 & 0.05 \\
        FCtL \cite{li2021contexts} & VGG-16 & Local & 72.87 & 3167 & 0.04 \\
        ISDNet \cite{guo2022isdnet} & ResNet-18 & Global & 74.23 & 4680 & 6.90 \\
        Ours & ResNet-18 & Global & \textbf{77.27} & 4834 & 5.53 \\
        \bottomrule
    \end{tabular}
    \label{tab:com_inriav2}
    \vspace{-0.3cm}
\end{table}

\begin{table}
    \centering
    \setlength{\tabcolsep}{3pt}
    \vspace{-0.3cm}
    \caption{Comparison with state-of-the-arts with large GPU memory limits on DeepGlobe. * represents our implementation.}
    \vspace{-0.3cm}
    \footnotesize
    \begin{tabular}{cccccc}
        \toprule
        Model & Backbone & Inference & mIoU & Memory & FPS \\
        \midrule
        U-Net* \cite{ronneberger2015u} & U-Net & Global & 28.53 & 3511 & 9.34 \\
        FCN-8s* \cite{long2015fully} & ResNet-18 & Global & 68.67 & 1890 & 7.98 \\
        DeepLab v3+* \cite{chen2018encoder} & ResNet-18 & Global & 72.18 & 2398 & 7.22 \\
        BiSeNetV1 \cite{yu2018bisenet} & ResNet-18 & Global & 53.00 & 1801 & 14.20 \\
        STDC  \cite{fan2021rethinking} & STDC & Global & 70.30 & 2580 & 14.00 \\
        PointRend \cite{kirillov2020pointrend} & ResNet-50 & Global & 71.78 & 1593 & 6.25 \\
        CascadePSP \cite{cheng2020cascadepsp} & ResNet-50 & Local & 68.50 & 3236 & 0.11 \\
        GLNet \cite{chen2019collaborative} & ResNet-50 & Local & 71.60 & 1865 & 0.17 \\
        MagNet-Fast \cite{huynh2021progressive} & ResNet-50 & Local & 71.85 & 1559 & 3.40 \\
        MagNet \cite{huynh2021progressive} & ResNet-50 & Local & 72.96 & 1559 & 0.80 \\
        FCtL \cite{li2021contexts} & VGG-16 & Local & 72.76 & 3167 & 0.13 \\
        ISDNet \cite{guo2022isdnet} & ResNet-18 & Global & 73.30 & 1948 & 27.70 \\
        Ours & ResNet-18 & Global & \textbf{73.50} & 1966 & 24.33 \\
        \bottomrule
    \end{tabular}
    \vspace{-0.3cm}
    \label{tab:com_deepglobev2}
\end{table}

\textbf{Memory-based interaction scheme.} In our memory scheme, we use semantic branch features to update the semantic memory, dubbed cross-branch interaction strategy. To demonstrate the necessity of this strategy, we perform ablation experiments with regard to it. Table~\ref{tab:ab_memory} shows performance barely changes (73.23\% v.s. 73.25\%) using the features of the memory-based branch, compared with no memory bank. For update mode, we ablate the effect of "Mean" mode (i.e., $\hat{\mathcal{R}}^c=\sum^{N^c}_{i=1}\frac{\mathcal{R}^c_i}{N^c}$). As shown in Table~\ref{tab:ab_memory}, this is only 0.06\% mIoU improvement over no memory bank. Besides, we replace Concat instead of Softmax (i.e., $M_b = \text{Concat}(\frac{\mathcal{M}^T \otimes \mathcal{F}_b}{\sqrt{D}}, M_b)$) for the reading mode, it leads to 0.08\% mIoU degradation. 

\begin{table}[t]
    \centering
    \setlength{\tabcolsep}{6pt}
    \vspace{-.3cm}
    \caption{Effectiveness of our proposed model structure.}
    \vspace{-0.3cm}
    \footnotesize
    \begin{tabular}{cccccc}
        \toprule
        Up-sampling & Spa. Inf. & Mem. & mIoU & Memory & FPS \\
        \midrule
        Bilinear & - & - & 71.81 & 1596 & 7.99 \\
        IFA~\cite{hu2022learning} & - & - & 72.23 & 1742 & 8.80 \\
        Ours & - & - & 69.24 & 1284 & 19.51 \\
        Ours & $M_b$ & - & 72.06 & 1470 & 10.56 \\
        Ours & $M_b$ & $\mathcal{M}$  & 72.53 & 1471 & 10.24 \\
        Ours & $M_b$+$M_l$ & - & 73.23 & 1471 & 10.13 \\
        Ours & $M_b$+$M_l$ & $\mathcal{M}$  & 73.66 & 1472 & 10.09 \\
        Ours & $M_b$+$M_l$ & $\mathcal{M}$+$\mathcal{M}_l$ & 73.09 & 1474 & 9.31 \\
        \bottomrule
    \end{tabular}
    \label{tab:ab_inr}
    \vspace{-.5cm}
\end{table}


\begin{table}
    \centering
    \setlength{\tabcolsep}{9pt}
    \caption{Different strategies for memory-based interaction scheme.}
    \vspace{-0.3cm}
    \footnotesize
    \begin{tabular}{ccccc}
        \toprule
        $\mathcal{M}$ & Cross-Branch & Update & Read & mIoU \\
        \midrule
        × & × & × & × & 73.23 \\
        \checkmark & × & Cosine & Softmax & 73.25 \\
        \checkmark & \checkmark & Mean  & Softmax & 73.29 \\
        \checkmark & \checkmark & Cosine & Concat & 73.15 \\
        \checkmark & \checkmark & Cosine & Softmax & 73.66 \\
        \bottomrule
    \end{tabular}
    \label{tab:ab_memory}
    \vspace{-.5cm}
\end{table}

\begin{figure}[t]
    \centering
    \vspace{-.5cm}
        \includegraphics[width=0.46\textwidth]{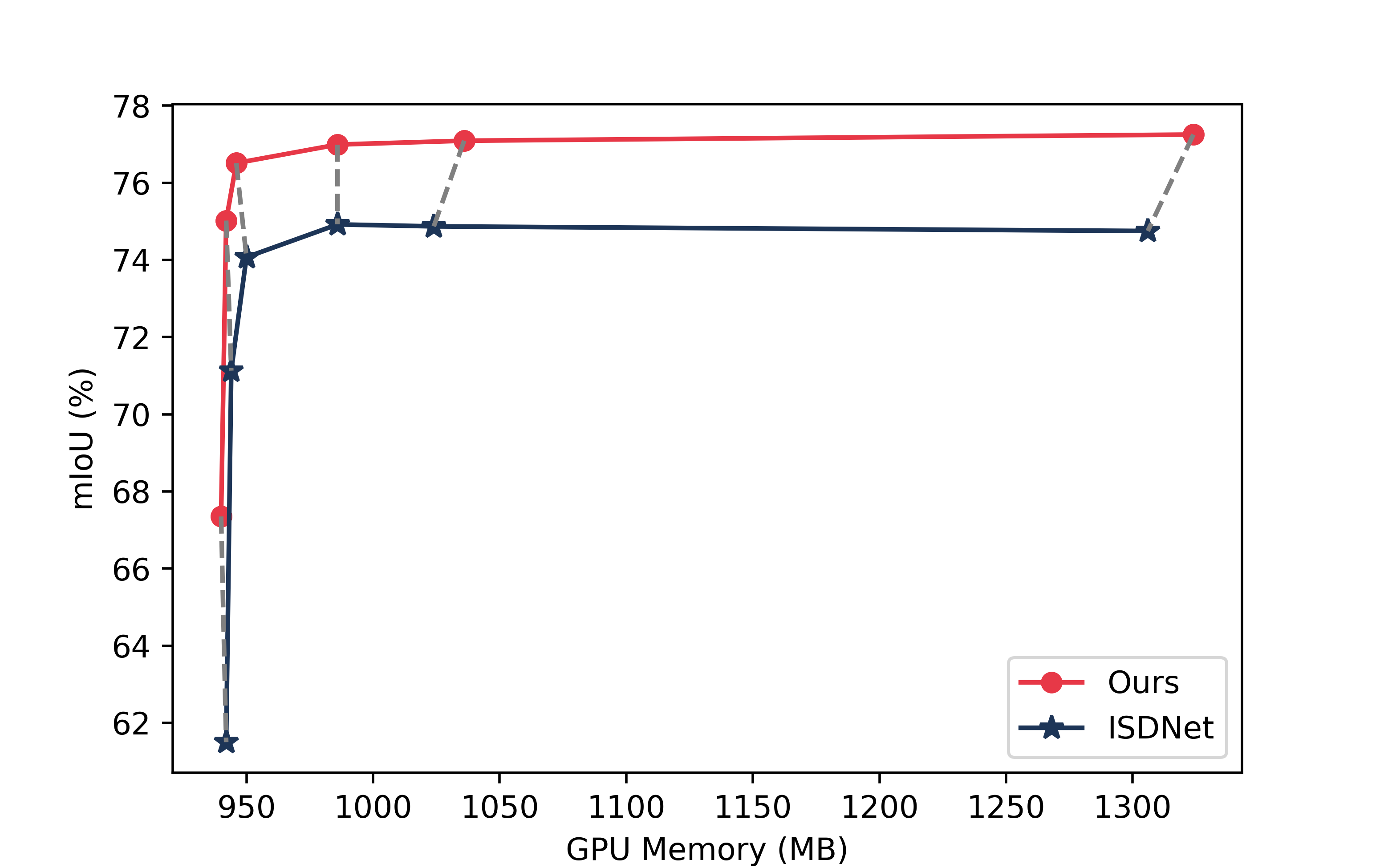} \\
        \vspace{-.3cm}
    \caption{By continuously reducing the size of local patches, we explore the minimum GPU memory cost of ISDNet and our approach pertaining to their corresponding segmentation performance.}
    \vspace{-.5cm}
    \label{fig:less}
\end{figure}

\textbf{Memory usage boundary.} \qi{To investigate the GPU memory usage boundary, we attempt to reduce the size of local patches in order to further decrease GPU memory consumption during the inference process. We perform this experiment on Inria Aerial with 5000×5000 pixel images. As shown in Fig.~\ref{fig:less}, our model and ISDNet achieve similar minimum GPU memory usage of approximately 940MB. However, even under the lowest-bound of GPU memory constraint, our model outperforms ISDNet by a significant margin, with comparatively less degradation in performance. In general, our model demonstrates superior performance over ISDNet, both under small and large GPU memory constraints.}

\section{Conclusion and Limitations}

In this paper, we propose an effective solution segmenting an ultra-high resolution image in a limited GPU memory system, which has practical value for robotic systems. In particular, we propose a spatial-guided high-resolution query module for local inference. Additionally, we also present a memory-based interaction scheme that efficiently enhances the semantics of high-resolution information by bridging cross-image contextual information.
There are several limitations in our method. First, it can hardly eliminate the noise from high-resolution information, leading to bias in the query process. Moreover, how to deploy the memory-efficient model while preserving high FPS remains a big challenge.

\bibliography{egbib}{}
\bibliographystyle{IEEEtran}

\end{document}